\documentclass[letterpaper,journal]{IEEEtran}
\usepackage{amsmath,amsfonts}
\usepackage{algorithm}
\usepackage{algorithmicx}
\usepackage{algpseudocode}
\usepackage[hidelinks]{hyperref}
\usepackage{array}
\usepackage[caption=false,font=normalsize,labelfont=sf,textfont=sf]{subfig}
\usepackage{lipsum}
\usepackage{textcomp}
\usepackage{stfloats}
\usepackage{url}
\usepackage{verbatim}
\usepackage{graphicx}
\usepackage{setspace}
\usepackage{tcolorbox}
\usepackage{cleveref}
\usepackage{physics}
\usepackage{multirow}
\usepackage{cite}
\usepackage{pifont}
\definecolor{darkblue}{RGB}{0,0,139}
\newcommand{\blue}{\textcolor{darkblue}}

\begin{document}

\title{Linear Mode Connectivity under Data Shifts for Deep Ensembles of Image Classifiers}

\author{C. Hepburn$^{*1}$, T. Zielke$^{*2}$ and A.P. Raulf$^{*3}$\\
Institute for AI Safety \& Security, German Aerospace Center (DLR)\\
\footnotesize{$^1$ https://orcid.org/0000-0002-2952-1729, 
$^2$ https://orcid.org/0009-0009-5164-7713, 
$^3$ https://orcid.org/0009-0003-8672-3014}
}

\maketitle

\begin{abstract}
The phenomenon of linear mode connectivity (LMC) links several aspects of deep learning, including training stability under noisy stochastic gradients, the smoothness and generalization of local minima (basins), the similarity and functional diversity of sampled models, and architectural effects on data processing. In this work, we experimentally study LMC under data shifts and identify conditions that mitigate their impact. We interpret data shifts as an additional source of stochastic gradient noise, which can be reduced through small learning rates and large batch sizes. These parameters influence whether models converge to the same local minimum or to regions of the loss landscape with varying smoothness and generalization. Although models sampled via LMC tend to make similar errors more frequently than those converging to different basins, the benefit of LMC lies in balancing training efficiency against the gains achieved from larger, more diverse ensembles. 
\normalfont{Code and supplementary materials are available at \href{https://github.com/DLR-KI/LMC}{https://github.com/DLR-KI/LMC}. This work has been submitted to the IEEE for possible publication. Copyright may be transferred without notice, after which this version may no longer be accessible.}
\end{abstract}


\section{Introduction}
\IEEEPARstart{M}{ode} connectivity refers to a phenomenon, when stochastic gradient descent (SGD) solutions or modes are connected via a path of low loss in neural networks parameter space \cite{garipov}, \cite{draxler}. So every solution along such path exhibits similar performance and generalization as those solutions, between which the path is constructed. Moreover, such paths were shown to be embedded in a multi-dimensional manifold of low loss \cite{benton}. When a connecting path is \emph{linear} the phenomenon is referred to as \emph{linear mode connectivity} (LMC) \cite{frankle}. LMC was investigated under different perspectives: (1) conditions affecting LMC \cite{frankle}, \cite{alt}, (2) connectivity of layers, features or different types of solutions \cite{llfc}, \cite{asja}, \cite{multitask} and (3) so-called "re-basin" approaches, that "transport" a solution from one local minimum\footnote{\emph{Local minimum} or \emph{basin} are used interchangeably throughout the paper.} (basin) to another \cite{ainsworth}, \cite{sinkhorn}, \cite{repair}, meaning that two solutions from the same basin are linearly connected \cite{entezari}. From a practical view point, LMC is expected to improve ensemble methods, in particular in federated learning setting, robustness of fine-tuned models, distributed optimization and model pruning \cite{wortsman}, \cite{ainsworth}. This work focuses on LMC from the perspective of \emph{data shifts} \cite{shifts}, which are ever-present in real world applications. In particular, when training is performed on multiple training datasets separately and ensembles of models are employed. We address the following questions:

    \begin{figure}[!ht]
        \centering
        \includegraphics[width=1\linewidth]{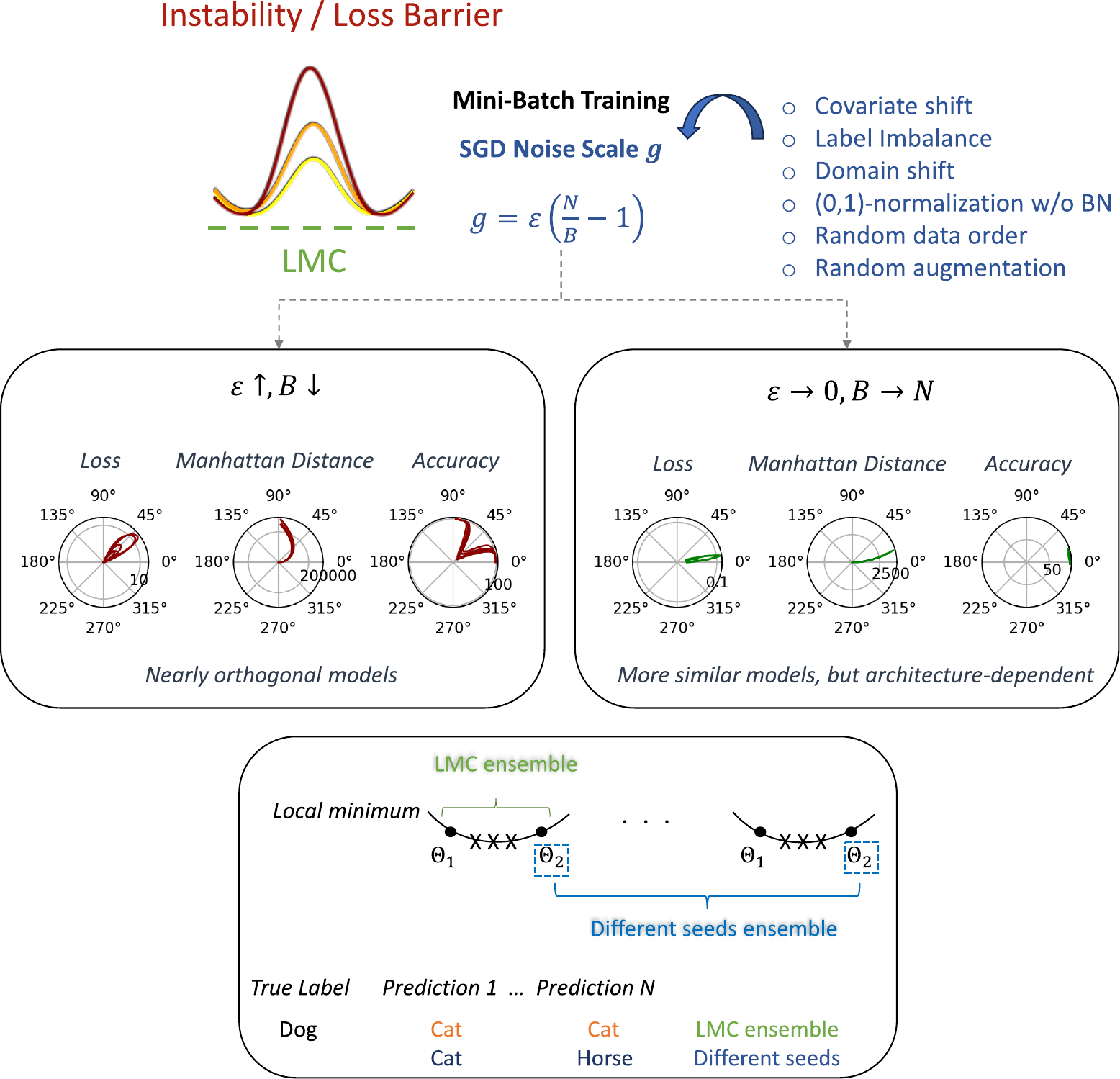}
        \caption{\textbf{The stability of mini-batch training depends on SGD noise scale $g$ \cite{smith}, a noisy estimate of the true loss gradient that arises from data shifts, as well as from random data sampling and/or augmentation \cite{frankle}. In large-batch or small-learning rate regimes \cite{alt}, this noise vanishes; these parameters influence convergence towards local minimum or regions of the loss landscape with varying smoothness and generalization properties \cite{visualizing}. Models, sampled via LMC tend to make the same mistake more often.}\\ Notation: w/o BN stands for 'without batch normalization'.}
        \label{conclusion}
    \end{figure}
    
\begin{enumerate}
    \item \emph{\textbf{Do data shifts result in instability of training dynamics, that is, do they alter optimization trajectory through loss landscape?}} Covariate shift, label imbalance and domain shift can "break" LMC. Interestingly, normalizing image intensities to the [0,1] range causes MLPs trained without batch normalization to converge to different local minima. Adding batch normalization, which reduces \emph{internal} covariate shift, layers restores stability in the training dynamics. However, linear interpolation plots may not always indicate true LMC: test loss can show a plateau caused by overfitting.
    \item \emph{\textbf{If instability is caused by data shifts, can it be mitigated by the size of a training mini-batch and a learning rate?}} The effects can be partly mitigated by increasing batch size or decreasing learning rate. Under instability, the loss barrier depends on these parameters; in some cases, barriers remain small (0.1-1), yet accuracy drops by up to 15\%. ResNets are most affected by label imbalance. Greater network depth can hinder the emergence of LMC.
    \item \emph{\textbf{What accounts for LMC "dependency" on training mini-batch size and learning rate and what are their effects on models?}} Increasing batch size and/or decreasing learning rate in general makes models more similar and "favors" LMC (\cref{conclusion}). MLPs can be nearly orthogonal and yet exhibit LMC, while deep models (ResNet, VGG) exhibit loss barrier, despite high similarity.\\
    Training stability depends on the scale of SGD noise $g$, which in mini-batch training arises from data shifts, random data order, and data augmentation. Stability is maintained when the learning rate $\varepsilon$ or the data size-to-batch ratio $\frac{N}{B}$ is small, causing models to follow similar trajectories through the loss landscape and converge to the same local minimum or region (i.e., exhibiting LMC and higher model similarity). Under instability, batch size and learning rate influence convergence towards regions of the loss landscape with different smoothness and, consequently, different loss barrier values. Moreover, large-batch training tends to favor sharper minima, which are associated with poorer generalization. Even within the same basin (small barrier values, i.e., minor loss fluctuations), sampled models can still exhibit up to a $15\%$ drop in performance.
    \item \emph{\textbf{What are the benefits of LMC for ensembling methods?}} We observed that models, sampled from the same basin, tend to make the same mistakes more often than models from different local minima (\cref{conclusion}). Since ensemble performance generally improves with ensemble size \cite{lobacheva}, the benefit of LMC-based sampling involves a trade-off between training many independent models and only a few that exhibit LMC.
\end{enumerate}

From a practical standpoint, our findings support the following recommendations, summarized in the \cref{suggestions}.

The paper is structured as follows. The Section II discusses the related work, the Section III describes the methodology (data, models, observables and metrics), the experimental results are presented and interpreted in the Section IV. The conclusion and future work are outlined in the Section V.

\section{Related work}
\subsection{Linear mode connectivity}
Frankle et al. experimentally investigated how stochastic gradient descent noise affects training "stability", where SGD noise arises from random mini-batch data order and augmentation\footnote{To elaborate, during each epoch the training examples in a mini-batch are fed into the network in random order, and may also be randomly augmented. Random data order and augmentation vary across training runs and can alter optimization trajectories despite identical initialization and hyper-parameters.} \cite{frankle}. Training two copies of the same network under different noise \emph{samples} fails to converge to the same local minimum with the exception of LeNet, trained on MNIST. Training remains stable, however, when two networks are further trained, initialized from the same pretrained weights. In contrast to this work, we fix SGD noise sample to insure that if instability occurs, it has data-related origin. Moreover, large-batch and small-learning rate training regimes, as well as optimizer choice, counteract the effects of SGD noise: Altintas et al. experimentally deduced the conditions in which LMC can be preserved for MLP models of varying depth, that have a correspondence with convolutional and attention-based models \cite{alt}. In our work, apart from different architecturs (MLP, VGG, ResNet), we were interested in small-batch training regime, that guides the covergence of SGD towards flat minima (\cref{loss_landscape}).
Additionally, small-batch training is beneficial for applications, that require high memory consumption \cite{GN}. Mirzadeh et al. investigated LMC in continual (sequential) and multitask (simultaneous) learning settings and observed that the solutions obtained in these settings are linearly connected \cite{multitask}. Entezari et al. studied LMC between solutions obtained from \emph{different} initializations, showing barrier dependency on network width and depth with deeper networks exhibiting much larger barriers \cite{entezari}. Furthermore LMC was observed in permutation-invariant networks. The authors of \cite{entezari} conjectured that once proper permutations of networks' parameters are found, the SGD solutions, that these networks represent, belong to the same basin. In accordance with this conjecture Ainsworth et al. proposed methods to find such permutations \cite{ainsworth}. Guerrero Pena et al. improved the method of \cite{ainsworth} by relaxing the constraint of permutation matrix to be binary, making the method differentiable and applicable to any loss objective \cite{sinkhorn}. Jordan et al. improved the neuron alignment method by reducing collapse in the variance of activations of the interpolated networks \cite{repair}.

\subsection{Parameterization, loss landscape \& generalization of neural networks}\label{loss_landscape}
Neural networks loss landscape is complex, exhibiting many (flat, sharp, asymmetric) local minima and saddle points \cite{haeffele}, \cite{asym}, \cite{visualizing}. Several works discussed the relation between sharpness / flatness of a local minimum and generalization ability of neural networks \cite{keskar} - \cite{neysh}. It was empirically demonstrated that training networks in a large-batch regime ($>$ 512 data samples) influences convergence of SGD towards \emph{sharp} minima, which are thought to be correlated with \emph{poor} generalization \cite{keskar}. Dinh et al., however, showed that typical definitions of flatness are not suitable measure of generalization ability of \emph{ReLU} networks due to their scale-invariance. A model can be re-parameterized, resulting in the \emph{same} function with \emph{different flatness} of a minimum \cite{dinh}. Finally, Li et al. proposed filter-wise normalization for random directions to visualize loss contours of \emph{rectified} networks. As a result, large batch training regime produces “visually”\footnote{As evidenced by linear interpolation plots (\cref{observables_metrics_section}).} sharper local minima, albeit to a limited extent, with poorer generalization \cite{visualizing}.
Essentially, parameterization of neural networks affects the number of minima and (non)-convexity of loss landscape. In simple terms, it depends on number of data samples $N$, data dimensionality $D$ and number of model parameters $P$ \cite{schaeffer}, \cite{nakkiran}, \cite{ascoli}. In contrast to over-parameterized models $P > N$, under-parameterized have less parameters than training data $P < N$, while networks in the interpolation threshold regime $P=N$ may perfectly fit the training data \cite{schaeffer}.

\begin{table*}[!ht]
    \renewcommand{\arraystretch}{1.2}
    \resizebox{\linewidth}{!}{
    \begin{tabular}{|c|c|c|c|c|c|c|c|}\hline\hline
        Ensemble & Training Models & Functional Diversity & LR & Batch Size & \begin{tabular}{c}
            Mini-Batch Training\\
            Stability Under *
        \end{tabular} & Minimum & Generalization \\\hline\hline
        LMC & 2 & Reduced & $\leq 10^{-3}$ & $\leq 512$ \cite{keskar}, \cite{smith} & \begin{tabular}{l}
        Stable** or re-basin\\
        \cite{ainsworth}, \cite{sinkhorn}, \cite{benedikt}
        \end{tabular} & Flat \cite{visualizing} & Better \cite{visualizing}\\\cline{4-8}
         &  &  & $\leq 10^{-3}$ & $> 512$ & Stable & Sharp & Poorer\\\hline
         Different Seeds & $N$ & Greater & *** & $\leq 512$ & - & Flat & Better\\\hline\hline
    \end{tabular}
    }
    \caption{\textbf{Advantages and limitations of deep ensembles sampled via LMC or trained with different seeds.}\\
    * data shifts, random data sampling and/or augmentation.\\
    ** ResNets are most affected by label imbalance; greater network depth can hinder the emergence of LMC.\\
    *** as required.}\label{suggestions}
\end{table*}

\section{Methodology}
\subsection{Data for Image Classification}
We consider three datasets for image classification, namely, MNIST \cite{mnist}, CIFAR-10 \cite{cifar} and BigEarthNet \cite{bigearth}. While CIFAR-10 small-scale dataset is far more heterogeneous than MNIST in terms of image content, patterns, background, BigEarthNet provides multi-spectral remote sensing imagery (\cref{scheme}). This large-scale dataset is composed of satellite image tiles of land cover, which was probed at light frequencies, ranging from ultra-blue to short wave infrared (443 nm to 2190 nm in central wavelength; appendix, \cref{bigearth_data}). 12 spectral bands result in multiple images per scene of variable spatial resolution and matrix sizes (120 $\times$ 120 pixels, 10 m spatial resolution; 60 $\times$ 60 pixels, 20 m; 20 $\times$ 20 pixels, 60 m). Images are non-overlapping with 95$\%$ having at most 5 labels per image. Data was acquired in ten countries of Europe (Austria, Belgium, Finland, Ireland, Kosovo, Lithuania, Luxembourg, Portugal, Serbia, Switzerland) and across all four seasons. Images with seasonal snow, cloud or cloud shadows were excluded from the training data.

\subsection{Data shifts}
\begin{figure}[!ht]
    \centering
    \includegraphics[width=0.8\linewidth]{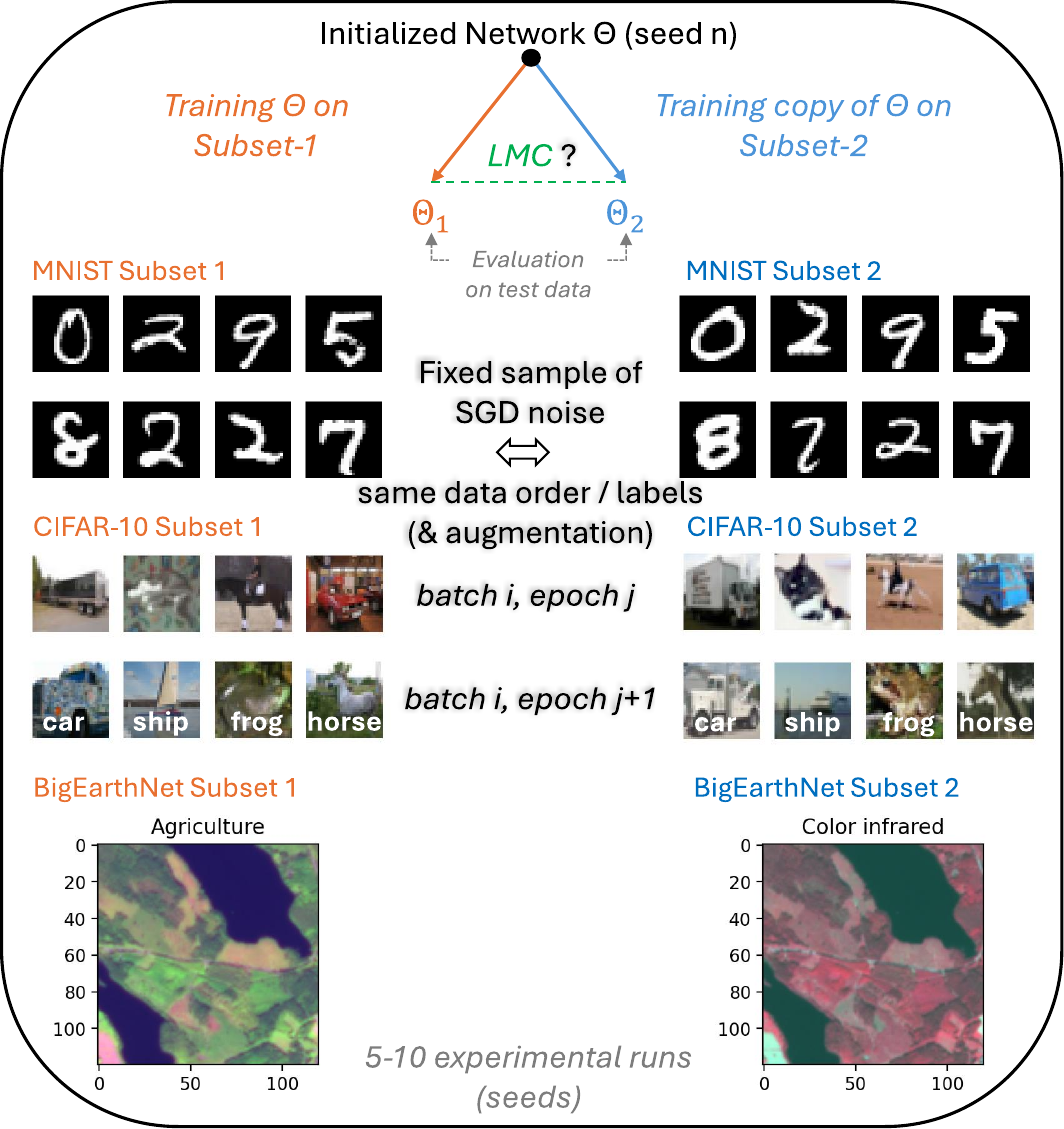}
    \caption{\textbf{Illustration of the training scheme.} Training data is partitioned into two disjoint subsets (\cref{partitions}). In case when SGD noise sample is fixed (as shown in the figure), \emph{image semantics is the same for the same batch and epoch across the training subsets}, but varies across epochs due to data shuffling (pseudo-code is presented in the Appendix, \cref{pseudocode}).}
    \label{scheme}
\end{figure}

To introduce \emph{data shifts}, we randomly split the training data into two subsets. The datasets and their partitions are summarized in the \cref{partitions}. We then initialize a network $\Theta$ and train two copies of it, each on one subset, under the same sample of SGD noise (unless noted otherwise), as schematically illustrated in the \cref{scheme}. While models are trained on disjoint subsets, their performance and performance of linearly interpolated models (\cref{observables_metrics_section}) is evaluated on \emph{test data}. That is, shift occurs between training subsets (hence possible alteration of training trajectory) and between training and test data (affects generalization).

\begin{itemize}
    \item \textbf{Label imbalance and covariate shift}: we split MNIST and CIFAR-10 training data in half at random with either the same or different percentage of labels in each subset, the latter introduces label imbalance. In case of disjoint subsets with equal amount of labels \emph{we assume} a covariate shift (change in image distribution) to some degree due to "shift" in image features. This could be, for example, a specific hand-writing style (slant to the right; MNIST) or a certain animal breed, background etc. (CIFAR-10), predominant in one training subset only. Test data is balanced in terms of labels.
    \item \textbf{Domain shift}: two subsets of BigEarthNet contain images from different bands or different combinations of bands. Each band provides images of the same scenes captured at different light frequencies. Domain shift arises from these changes in measurement conditions. In case of combination of bands, each combination emphasizes different image features, such as healthy or unhealthy vegetation, denser vegetation or urban areas\footnote{ \href{https://gisgeography.com/sentinel-2-bands-combinations/}{https://gisgeography.com/sentinel-2-bands-combinations/}}. Test data contains both bands or both combinations of bands\footnote{For example, when one subset contains images captured at the blue light frequency and the other at the green, the test set contains images captured at both blue and green.}.
\end{itemize}

\subsection{MLP, VGG and Resnet Models}
We consider multi-layered perceptrons (MLPs), VGG \cite{vgg} and ResNet \cite{resnet} models of variable depth in over-parameterized regime (Appendix, \cref{models}). Each architecture processes data differently, and therefore the observed effects can be either model-specific or model-agnostic. By processing image data as input vectors, MLP ignores spatial structure and treats each pixel independently. In contrast, VGG models contain convolutional layers that capture local patterns and hierarchal features. ResNet models employ both convolutional layers and residual skip connections, which help to preserve low-level information. Details on model architectures and training hyper-parameters can be found in the supplementary material. For each experimental setting, we perform (mostly) five runs that differ in initialization seed of model parameters.

\begin{table*}[!htt]
    \resizebox{\linewidth}{!}{
    \begin{tabular}{|c|c|c|c|c|c|}\hline\hline
        Dataset & Size & Classes & Class / Image & \textbf{Training Data Partition} & Comments  \\\hline
        MNIST & 70,000 & 10 & 1 &
        \begin{tabular}{lcl}
            subset-1 & x & $\%$ of labels 0-4\\
            &(100-x) & $\%$ of labels 5-9\\
            subset-2 & (100-x)& $\%$ of labels 0-4\\
            & x & $\%$ of labels 5-9
        \end{tabular} &
        \begin{tabular}{ll}
            $x=50\%$ & (split 50/50)\\
            &- half of training data\\
            &- disjoint subsets\\
            &- \textbf{covariate shift}\\
            &- \textbf{fixed SGD noise}\\
            $x=20\%$ & (split 20/80)\\
            &- half of training data\\
            &- disjoint subsets\\
            &- \textbf{covariate shift}\\
            &- \textbf{label imbalance}\\
            &- \textbf{SGD noise}\\
        \end{tabular}\\
        CIFAR & 60,000 & 10 & 1 &  &\\\hline
        BigEarthNet & 480,038 & 19 & up to 5 & 
        \begin{tabular}{ll}
            subset-1 & band $B_i$ (or $B_iB_jB_k$)\\
            subset-2 & band $B_j$ (or $B_lB_mB_n$)\\
        \end{tabular} &
        \begin{tabular}{l}
            - full dataset\\
            - \textbf{domain shift}\\
            - \textbf{fixed SGD noise}\\
        \end{tabular}\\
        \hline\hline
    \end{tabular}}
    \caption{Datasets and training data partitions.}
    \label{partitions}
\end{table*}

\subsection{Observables \& Metrics}\label{observables_metrics_section}

\subsubsection*{Linear Interpolation}
Given a loss function $\mathcal{L}$ and two models $\mathbf{\Theta}_A,~\mathbf{\Theta}_B$ (a vectorized representation of parameters of models A and B, respectively) linear interpolation is performed by taking a linear combination of the two models, $\mathbf{\Theta} = (1-\lambda) \mathbf{\Theta}_A\ + \lambda \mathbf{\Theta}_B$ and evaluating loss at this new point, $\mathcal{L}(\mathbf{\Theta})$ either on test or training data for the parameter $\lambda \in [0,1]$. Goodfellow et al. introduced it to study the behavior of loss function along a linear path in the parameter space \cite{goodfellow}.

\subsubsection*{Loss Barrier}
Multiple variations of loss barrier exist in the literature \cite{frankle}, \cite{entezari}, \cite{alt}. We consider the definition of \cite{frankle}, which specifies loss barrier $\mathcal{B}$ as the difference between maximal value of loss function on the linear interpolation between two models and average loss of the two models \cite{frankle}:

\begin{equation}
\begin{array}{l}
\mathcal{B}(\mathbf{\Theta}_A,\mathbf{\Theta}_B) =\\[2ex]
\max_\lambda \Big[\mathcal{L} \Big( (1-\lambda) \mathbf{\Theta}_A\ + \lambda \mathbf{\Theta}_B \Big)\Big] - \frac{1}{2}\Big[\mathcal{L}(\mathbf{\Theta}_A) + \mathcal{L}(\mathbf{\Theta}_B)\Big],\\[2ex]
~\lambda \in [0,1],
\end{array}
\end{equation}
Importantly, two SGD solutions, $\mathbf{\Theta}_A$ and $\mathbf{\Theta}_B$, exhibit LMC, that is, are in the same local minimum, if $ \mathcal{B}(\mathbf{\Theta}_A,\mathbf{\Theta}_B) \approx 0$ along a linear path.\\
Another definition of loss barrier assigns no value to a linear variation of loss with respect to the interpolation parameter \cite{entezari} and Altintas et al. proposed to normalize barrier value by average test accuracy \cite{alt}.

\subsubsection*{Accuracy difference}
Drop or gain in performance is measured by difference $\Delta$ in accuracy (or weighted average precision\footnote{ \href{https://scikit-learn.org/stable/modules/generated/sklearn.metrics.average_precision_score.html}{\scriptsize{https://scikit-learn.org/stable/modules/generated/sklearn.metrics.average\_precision\_score.html} } } in case of BigEarthNet) with respect to average accuracy (mean weighted average precision) of models, between which interpolation is performed: 

\begin{equation}
    \Delta = \text{Acc}\Big(f_{\mathbf{(1-\lambda) \mathbf{\Theta}_A + \lambda \mathbf{\Theta}_B}}\Big) - \frac{1}{2} \Big[\text{Acc}(f_{\mathbf{\Theta}_A}) + \text{Acc}(f_{\mathbf{\Theta}_B})\Big],\label{Delta}
\end{equation}
where $f_\mathbf{\Theta}$ is a function represented by a neural network with parameters $\mathbf{\Theta}$, Acc$\Big(f_{\mathbf{(1-\lambda) \mathbf{\Theta}_A + \lambda \mathbf{\Theta}_B}}\Big)$ is accuracy of an interpolated model, exhibiting loss barrier or a minimum (\cref{interp_curves}), as evaluated either on test or training data.

\subsubsection*{Similarity metrics}

Linear mode connectivity implies possible models’ similarity in terms of parameters,  representations and functions. To analyze the effects of large batch-size and small learning rate training, we use standard metrics such as \emph{cosine similarity, which is scale-invariant}, and \emph{Manhattan distance} (formal definitions are given in the appendix, \cref{sim_metrics}).

\subsubsection*{Average wrong agreement and disagreement}

To look into the question of whether models from the same basin make the same mistake more often than models from different basins, we compute (formal definitions in the appendix, \cref{wa_wd}):

\begin{itemize}
    \item \emph{Average wrong agreement, $\overline{WA} \in [0,1]$}: proportion of times, when models agree in their predictions and are wrong
\item \emph{Average wrong disagreement, $\overline{WD} \in [0,1]$}: proportion of times, when models disagree in their predictions and are wrong.
\end{itemize}
The question of functional diversity was investigated in the \cite{fort}. We found it more precise to assess functional diversity not by measuring how much models agree or disagree in their predictions, but rather by examining the extent to which they agree on their errors. In the context of deep ensembles, models' disagreement in prediction can cancel errors of individual models, which arise due to difficult or erroneous data samples. Therefore, it is of interest to have "diverse" classifiers in the sense that, some models give wrong predictions, others - correct, but in their majority (either by vote or by averaging prediction vectors) the final answer is correct. This means that the situation, where every two models agree or disagree on prediction and both are wrong\footnote{Toy example:\\[1.2ex]
\begin{tabular}{c|c|c|l}
    \emph{True label} & \emph{ Prediction 1} & \emph{Prediction 2} \\
    Dog & Dog & Cat & \\
    & \blue{Horse} & \blue{Cat} & Wrong Disagreement\\
    & \textcolor{orange}{Cat} & \textcolor{orange}{Cat} & Wrong Agreement\\
    & Dog & Dog & \\
\end{tabular}
}, is less favorable to build a "good" ensemble. Specifically, if models sampled through LMC \emph{consistently make the same mistakes}, the resulting ensemble will exhibit a systematic bias.

\section{Experimental Results and Discussion}

\subsection{Remark on Measured Loss Barriers}

\begin{figure}[!ht]
    \centering
    \includegraphics[width=1\linewidth]{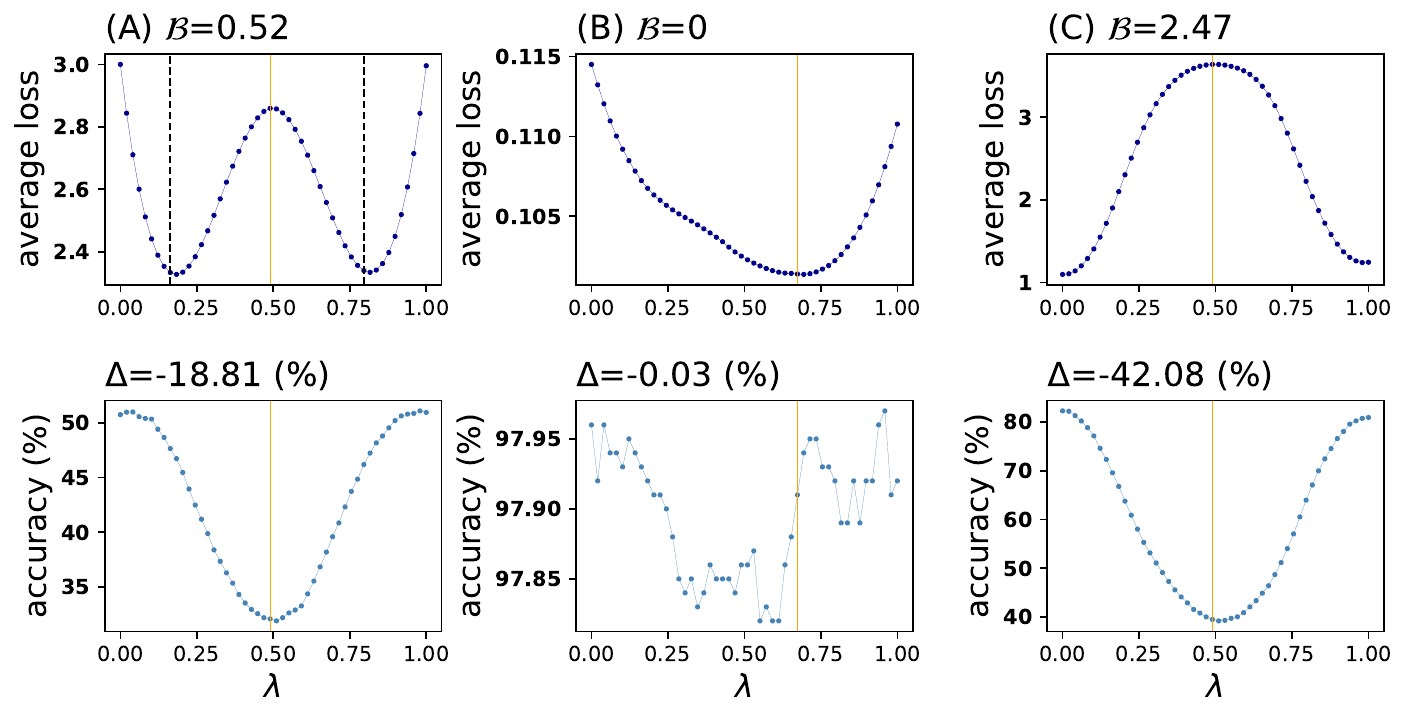}
    \caption{\textbf{Interpolated models may yield lower loss than the models they are interpolated between. When appropriate, barrier is measured with respect to local minima \emph{in "$\mathbf{\lambda}$}} (shown with dashed lines in A.)\\
    \textbf{Top}: average loss versus interpolation parameter $\lambda$ shows different "basin structures". \textbf{Bottom}: corresponding accuracy (test data). The vertical orange line indicates the value of $\lambda$, at which barrier, $\mathcal{B}$ and accuracy difference, $\Delta$ were measured.}
    \label{interp_curves}
\end{figure}

In our experiments, loss of \emph{some interpolated} models was sometimes lower than the loss of the models between which the interpolation was performed. However, such loss fluctuations exhibited a barrier with a corresponding drop in accuracy (\cref{{interp_curves}} A). To account for it, we measure barrier with respect to \emph{average local minimum in $\lambda$}. A convex-like "basin structure" assigns a zero value to the loss barrier (\cref{interp_curves} B).

\subsection{Covariate Shift Alters SGD Trajectories in the Small Batch Regime, Even Under a Fixed SGD Noise Sample}

\subsubsection*{Details}
We first trained two copies of the same model on disjoint data subsets with \emph{equal} amount of labels (possible covariate shift). We fix the SGD noise sample to isolate the effects of data shift on training dynamics, as schematically illustrated in \cref{scheme}. The models were trained in the small batch regime (batch size 32) with learning rates $10^{-3}$ or $10^{-4}$. The choice of batch size was motivated by the fact that small batch training "favors" flat minima with good generalization (\cref{loss_landscape}), which is beneficial for model ensembles.
\subsubsection*{Discussion}
\begin{figure}[!ht]
    \centering
    \includegraphics[width=1\linewidth]{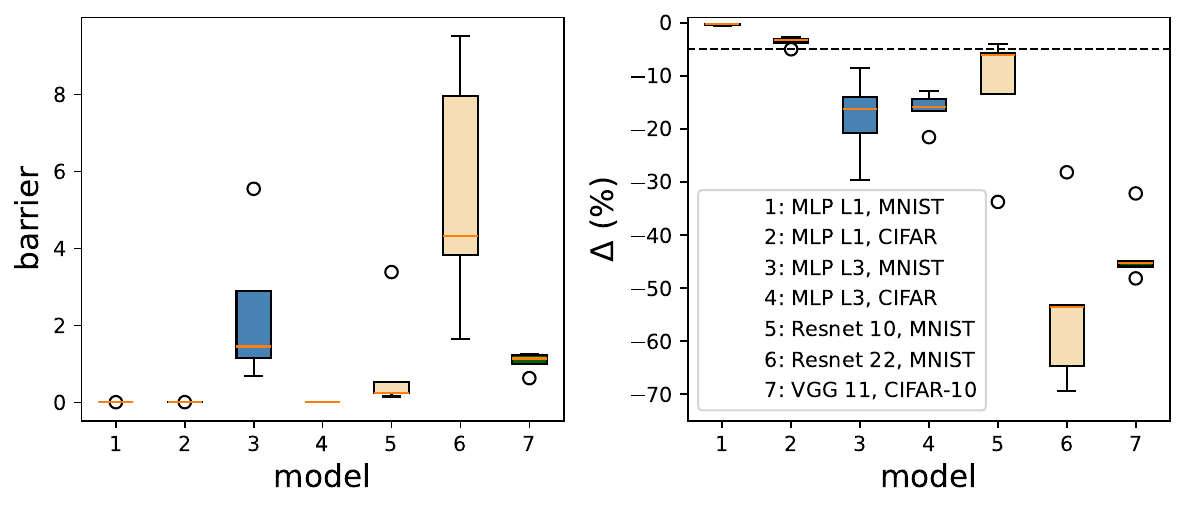}
    \caption{\textbf{Non-zero barriers and significant drop in accuracy of deep models suggest that covariate shift alone alters SGD training trajectories. One-hidden-layer MLPs have stable training dynamics.}\\
    Barrier, $\mathcal{B}$ and accuracy difference, $\Delta$ on linear interpolation for models of variable depth in small batch training regime (batch size 32) under a fixed SGD noise sample. Learning rates: $10^{-3}$ for MLP with one (L1) and three (L3) hidden layers and ResNet, $10^{-4}$  for VGG. Values from five experimental runs (different seeds). Horizontal dashed line indicates accuracy difference of $-5\%$.}
    \label{interp_50_50}
\end{figure}

With the exception of one-hidden-layer MLPs (using the same parameterization as deep MLPs), we observed loss barriers and significant performance drops in interpolated models (\cref{interp_50_50}).

The results suggest that \emph{covariate shift alone} is enough to alter the optimization trajectory causing the deep models to be in different basins of the loss landscape at the end of training.\\
Entezari et al. shown experimentally that in very wide networks the barrier becomes small on a linear interpolation between models, trained from \emph{different} initializations \cite{entezari}. Here we observed that one-hidden-layer MLPs, trained from the same initialization, do not exhibit barriers and a drop in accuracy of up to 5$\%$ indicate the stability of their training dynamics to covariate shift \emph{under the chosen conditions}.\\
We observe that for one model (MLP L3 on CIFAR) interpolated models exhibit a $\sim 16\%$ performance drop on average despite zero barriers. We attribute such effect to \emph{over-fitting}, as discussed in the \cref{overfitting_section}.

\subsection{Effects of Overfitting: Linear Interpolation Plots May Not Indicate True LMC}\label{overfitting_section}

\begin{figure}[!ht]
    \centering
    \includegraphics[width=1\linewidth]{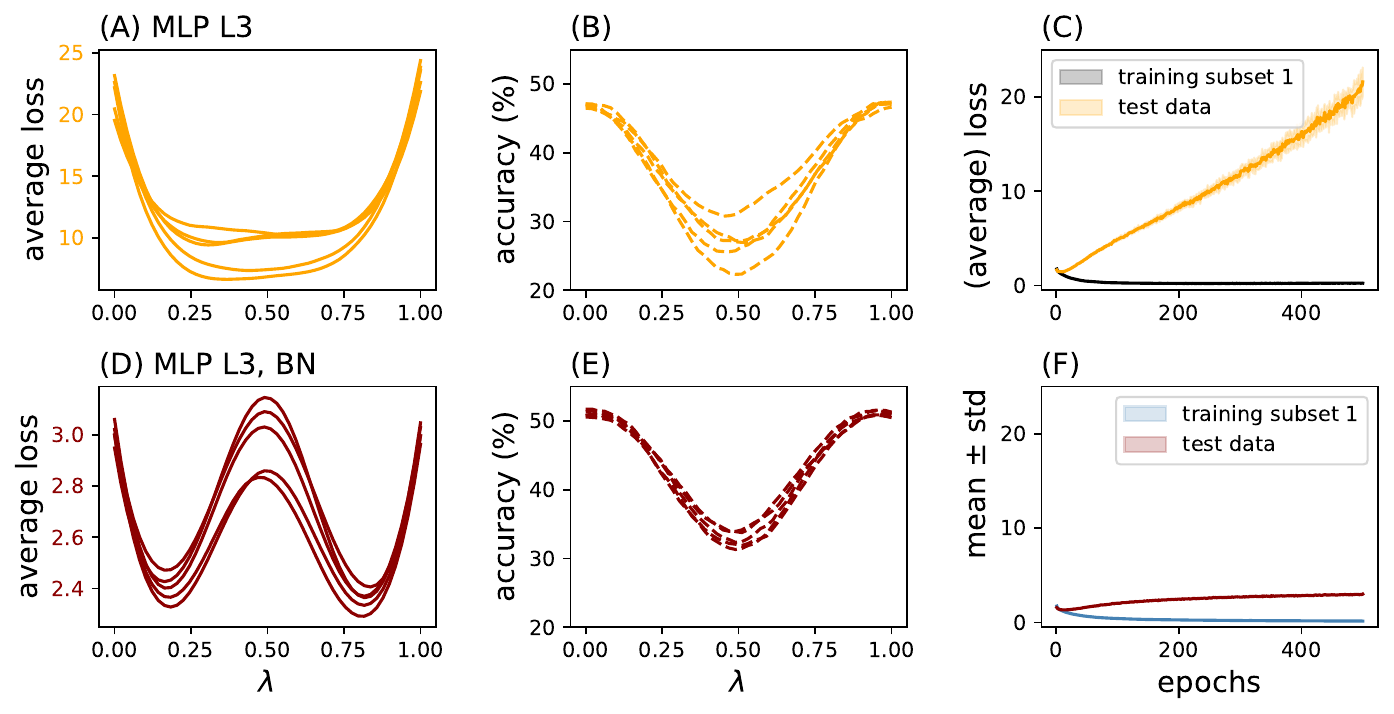}
    \caption{\textbf{Overfitting can explain poor generalization of interpolated models despite zero barriers.}\\
    (A)-(C): average learning curves and interpolation curves for three-hidden-layer MLPs, trained without batch normalization on CIFAR-10 subsets with equal label splits under a fixed SGD noise sample (five experimental runs, batch size 32, learning rate $10^{-3}$). Interpolated models are evaluated on test data.
    (D)-(F): same for MLPs, trained with batch normalization. Notation: MLP L3 three-hidden-layer MLP, BN batch normalization.}
    \label{overfitting}
\end{figure}

\begin{figure}[!ht]
    \centering
    \includegraphics[width=1\linewidth]{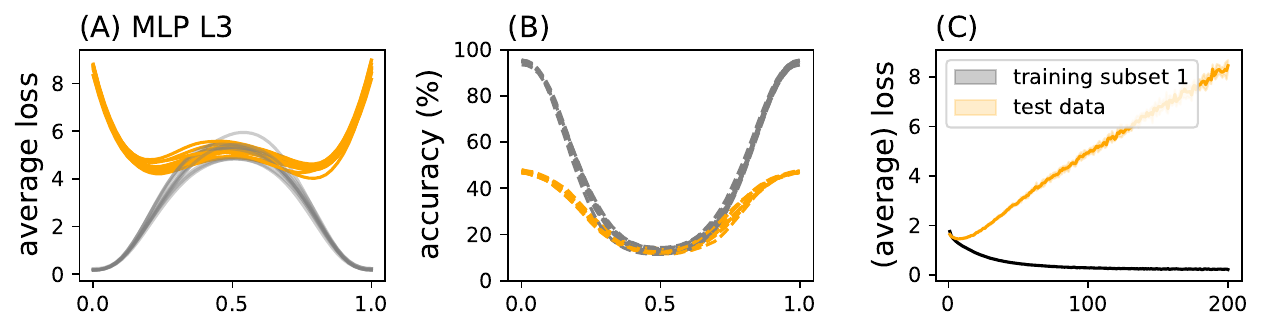}
    \caption{\textbf{Test losses are higher due to overfitting, with a \emph{a flattened area} corresponding to the training loss barrier.} Linear combination of the "end-points" models partially mitigates overfitting, lowering test loss; however, this is insufficient to improve generalization, as training accuracy drops by about 80\%. Models, trained from different initializations on the same CIFAR data subset (50/50 split) under different samples of SGD noise (batch size 32, learning rate $10^{-3}$).}
    \label{interp_eval_train_data}
\end{figure}

For three-hidden-layer MLPs trained on CIFAR-10 subsets with equal label splits, interpolated models show a $\sim 16\%$ performance drop on average despite zero barriers (\cref{overfitting} (A), (B)). We attribute this to \emph{overfitting}, as seen in the learning curves (\cref{overfitting} (C)). Linear interpolation reveals a convex-like basin yet linear combinations of models still overfit, leading to even poorer generalization. This interpretation is supported by batch normalization experiments: BN regularizes training, reduces overfitting, and yields lower final loss (\cref{overfitting} (F)). The presence of barriers and a $\sim 18\%$ performance drop on average in interpolated models suggests convergence to different local minima.\\
Note that apparent absence of barrier on linear interpolation between three-hidden-layer MLPs without batch normalization \emph{may not imply models' converged to the same local minimum.} The \cref{interp_eval_train_data} shows linear interpolation curves for models, trained from \emph{different initializations on the same data subset under different samples of SGD noise}. Interpolated models were evaluated on both training subsets and test data. On the training data, a barrier and corresponding accuracy drop appear along the linear path, indicating that the endpoint models converged to distinct local minima. On the test data, losses are higher due to overfitting, with a flattened area corresponding to the training loss barrier. A likely explanation is that linear interpolation between the endpoint models partially mitigates overfitting, lowering test loss; however, this is insufficient to improve generalization, as training accuracy drops by about 80\%.

\subsection{Large Batch Size / Small Learning Rate Training Regime Mitigates the Effects of Covariate Shift, "Recovering" LMC}\label{section_batch_lr}

\subsubsection*{Details}
As previously shown, covariate shift alone is enough to alter training trajectory. In these experiments, we train two copies of the same model on disjoint balanced data subsets under the same SGD noise samples, but in different batch / learning rate regimes. We investigate how such regimes affect stability of model training dynamics.

\subsubsection*{Observations}
\begin{figure}
    \centering
    \includegraphics[width=1\linewidth]{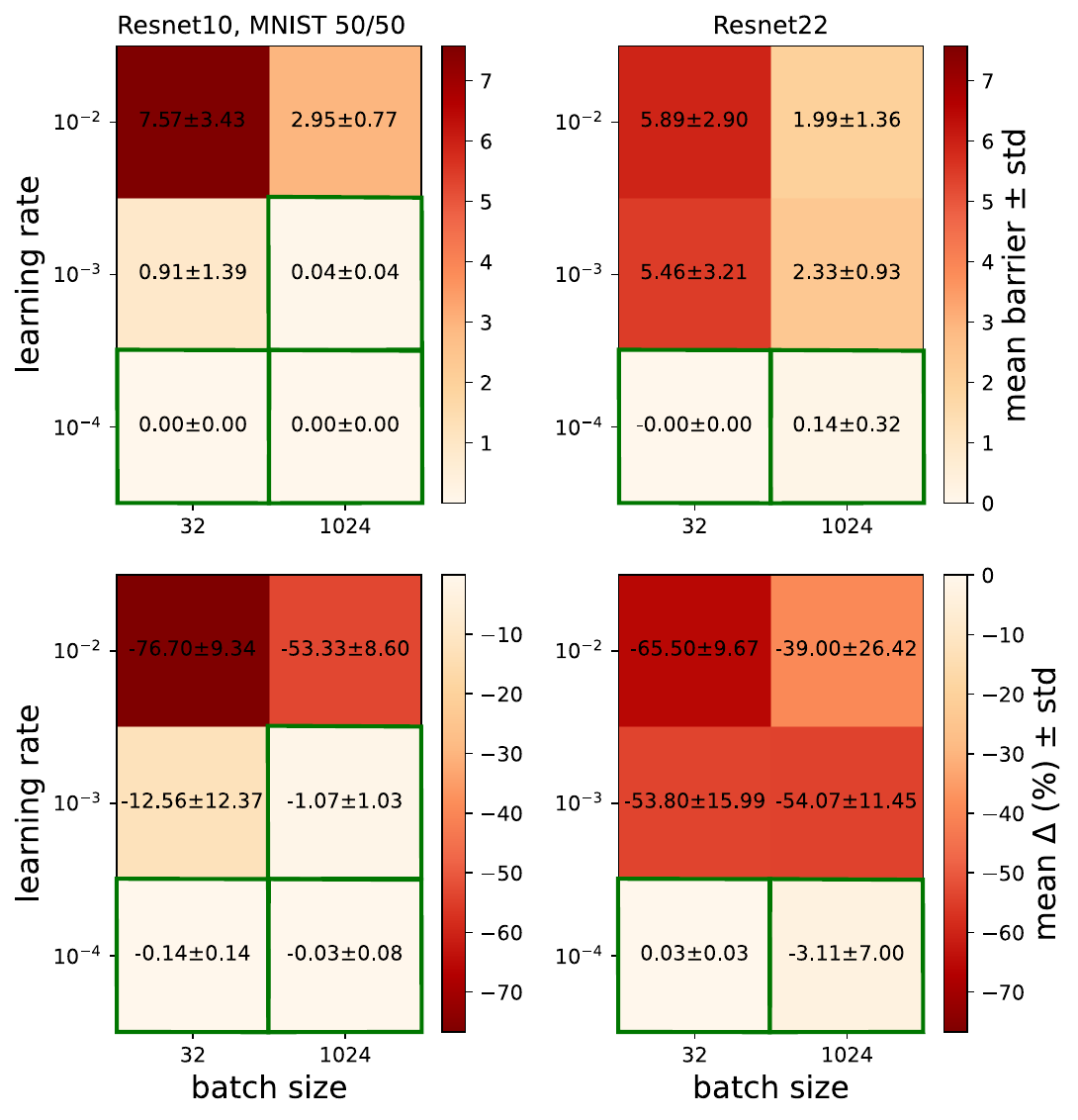}
    \caption{\textbf{Large batch size and small learning rate regimes mitigate the effects of covariate shift, "recovering" LMC.}\\
    Barrier, $\mathcal{B}$ (\textbf{top}) and accuracy difference, $\Delta$ (\textbf{bottom}) for ResNet models of variable depth, trained on disjoint subsets of MNIST with equal amount of labels under a fixed sample of SGD noise and variable conditions (batch size, learning rate).}
    \label{batch_lr_tradeoff}
\end{figure}

Training in large batch and/or small learning rate regime mitigates the effects of covariate shift (and of SGD noise, as shown in the \cite{alt}), resulting in models converging to the same local minimum. For example, the \cref{batch_lr_tradeoff} shows that for ResNet models, trained on disjoint MNIST subsets with the equal number of labels, the \emph{barrier decreases} as (1) batch size \emph{increases} for a fixed learning rate or (2) learning rate \emph{decreases} for a fixed batch size, and the models eventually exhibit LMC. Deeper ResNets lead to greater instability in training dynamics under the same training conditions. This trend is observed for wide, deep MLPs and VGG models both under covariate shift and different SGD noise samples (the results are presented in the appendix \cref{tables}, table VI).\\
Note that, in the presence of instability we observe a dependency of the barrier on learning rate and batch size. The observations presented are discussed in the \cref{section_bs_lr_effects}.

\subsection{ResNet training dynamics are more unstable under label imbalance compared to MLPs and VGG}

\subsubsection*{Details}
As shown in the previous section, choosing "the right" batch size and learning rate mitigates the effects of covariate shift, rendering training stable (the emergence of LMC). We chose such training conditions to investigate the effects of label imbalance, and domain shift (the latter is described in the \cref{domain_shift_section}). Here, we trained two copies of the same model on disjoint data subsets with a 20/80 label imbalance, which consequently introduced covariate shift and SGD noise.

\newpage
\subsubsection*{Discussion}
\begin{figure}[!ht]
    \centering
    \includegraphics[width=1\linewidth]{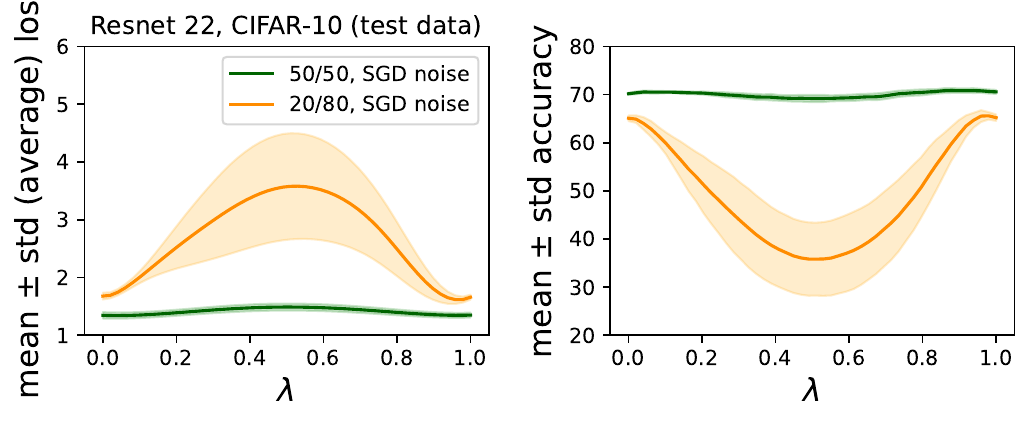}
    \caption{\textbf{ResNet training dynamics are unstable under label imbalance, which additionally introduces covariate shift and SGD noise. This may be linked to model's ability to preserve low-level information through residual skip connections.}\\
    Average loss (left) and accuracy (right) of interpolated models from five experimental runs (mean curves with standard deviation (shaded)). Linear interpolation was performed between models, trained on disjoint data subsets with equal (50/50) and imbalanced (20/80) label distributions additionally \emph{ under different SGD noise samples}.}
    \label{resnet_label_imbalance}
\end{figure}

\begin{figure}[!ht]
    \centering
    \includegraphics[width=1\linewidth]{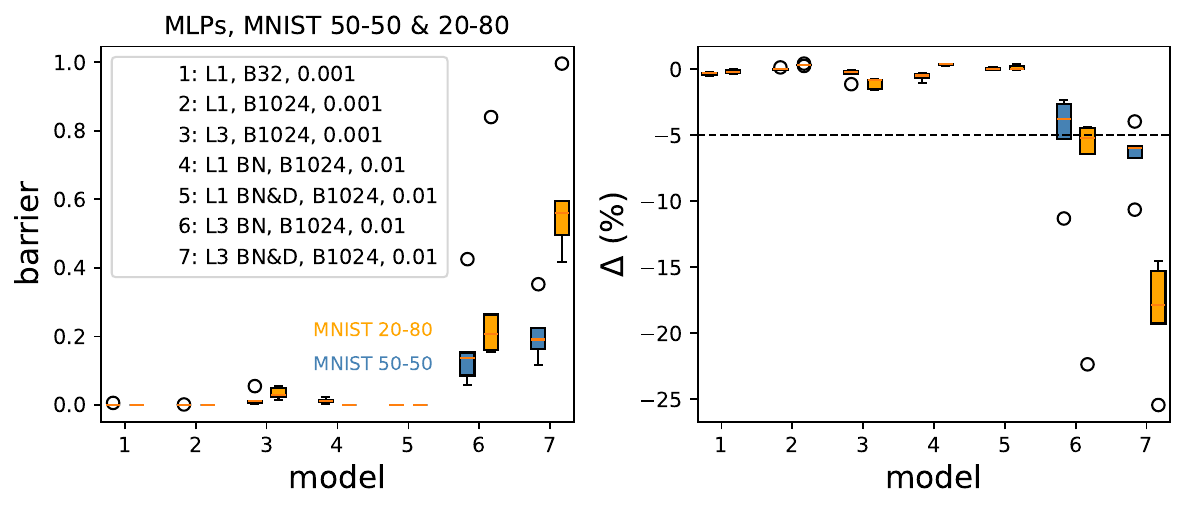}
    \includegraphics[width=1\linewidth]{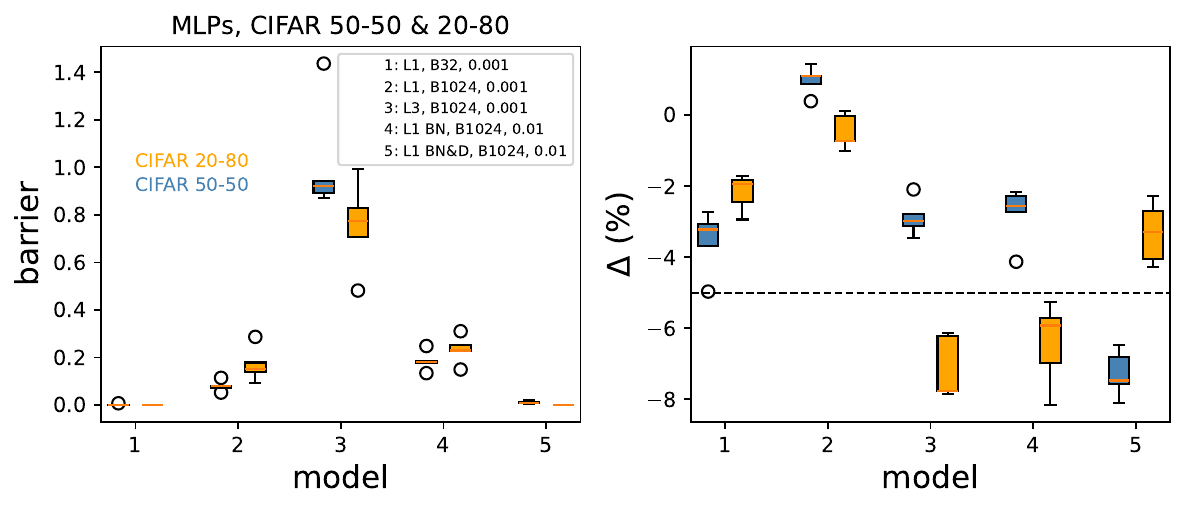}
    \caption{\textbf{MLP models exhibit stable training dynamics under label imbalance, sometimes with poorer generalization of the interpolated models.}\\
    Variable MLP models, trained on disjoint data subsets with equal (50/50) and imbalanced (20/80) label distributions. Batch size (B) and learning rates values are indicated in the figure insets. \textbf{Top:} MNIST data, \textbf{bottom}: CIFAR-10 data. Notation: BN stands for batch normalization, BN\&D stands for batch normalization followed by dropout.}
    \label{mlp_interp}
\end{figure}

Our results indicate that ResNet models are the most affected by label imbalance (Appendix \cref{tables}, table VII):

\begin{itemize}
    \item[\ding{118}] \textbf{MLP}: MLP models exhibited stable training dynamics under label imbalance (as well as under covariate shift and SGD noise), sometimes with poorer generalization of the interpolated models (Appendix, \cref{mlp_interp}).
    \item[\ding{118}] \textbf{VGG:} VGG11 remained stable under label imbalance. However, with increasing depth, stability is lost.
    \item[\ding{118}] \textbf{ResNet}: training ResNets was unstable under label imbalance, although stability was achieved when training on balanced subsets (50/50 split) \emph{with different samples of SGD noise} (\cref{resnet_label_imbalance}). 
\end{itemize}

Instability of ResNet training under label imbalance may be linked to their ability to preserve low-level information through residual skip connections. This resembles the behavior of VGG, where instability increases with depth: deeper networks capture more specific features (introduced here by label imbalance) in high-level representations. The ResNet results further suggest that similar instability could arise in architectures such as UNet \cite{unet}, which also use skip connections to combine low- and high-level features.

\subsection{Larger Domain Shifts Worsen Generalization of Interpolated Models Despite Small Barriers}\label{domain_shift_section}

\subsubsection*{Details}
We trained VGG-like models on BigEarthNet subsets, containing images from different bands or different band combinations (\cref{partitions}) under a fixed SGD noise sample (\cref{scheme}). We set the learning rate at $10^{-5}$ and the batch size at 256, since a larger batch and a smaller learning rate appeared to guide convergence to the same local minimum.

\subsubsection*{Discussion}
\begin{figure}[!ht]
    \centering
    \includegraphics[width=1\linewidth]{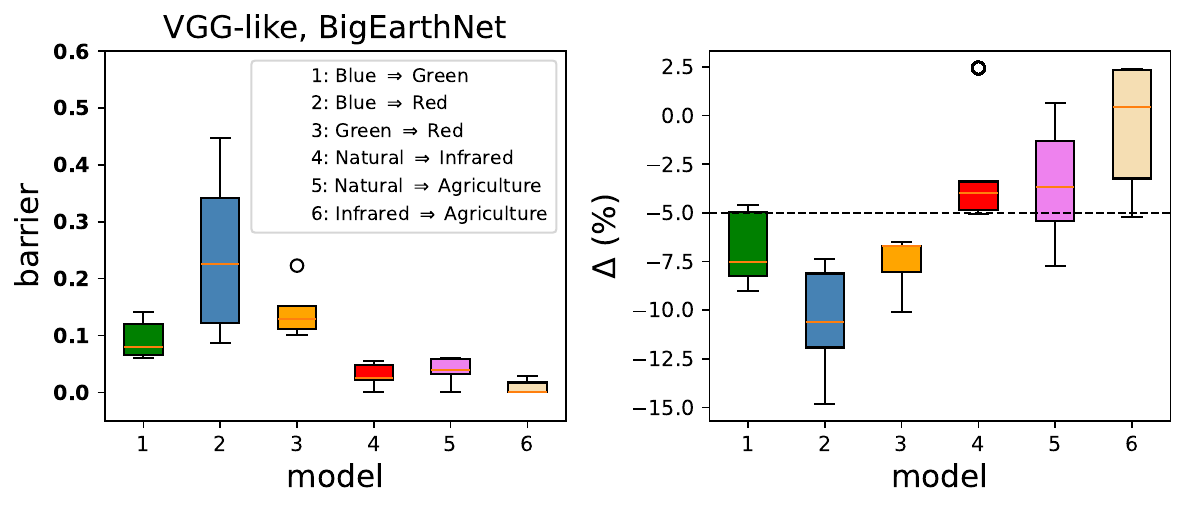}
    \caption{\textbf{Measured barriers may be small, yet interpolated models can lose up to $15\%$ accuracy.}\\
    Results of linear interpolation between models, trained on different bands (or band combinations; \cref{scheme}) and evaluated on test data, containing both. $\Delta$ describes difference in \emph{weighted average precision}.}
    \label{interpolations_bigearth}
\end{figure}

The interpolations show very small or near-zero barrier values (\cref{interpolations_bigearth}). However, generalization of interpolated models worsens when the models are trained on single bands. This effect is especially pronounced when the domain shift is larger, i.e., when the difference in probing light frequency is greater. We discuss this point in the \cref{section_bs_lr_effects}.\\
When training on band combinations, one or two bands overlap between combinations\footnote{For example, natural color combination contains B4, B3 and B2 bands, while color infrared combination includes B8, B4, B3.}, resulting in shared image features. This can explain near-zero barriers and smaller drops in weighted average precision of the interpolated models.

\subsection{Normalizing Image Intensities to the [0,1] Range "Breaks" LMC in MLPs, but Batch Normalization "Recovers" it}\label{norm_section}

\subsubsection*{Details}
\begin{figure}[!ht]
    \centering
    \includegraphics[width=0.45\linewidth]{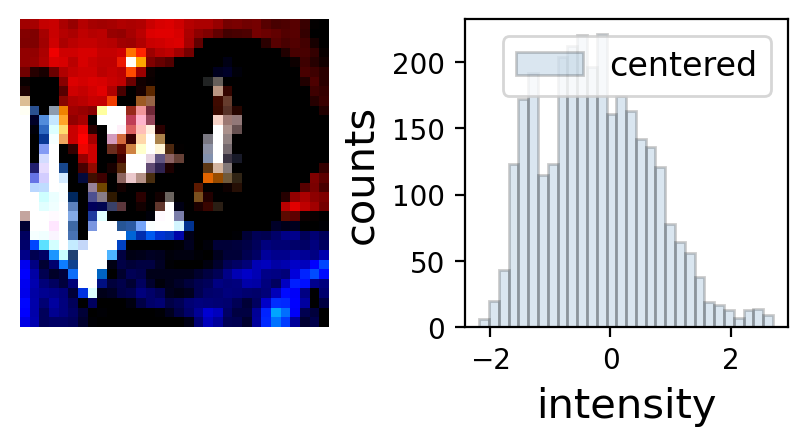}
    \includegraphics[width=0.45\linewidth]{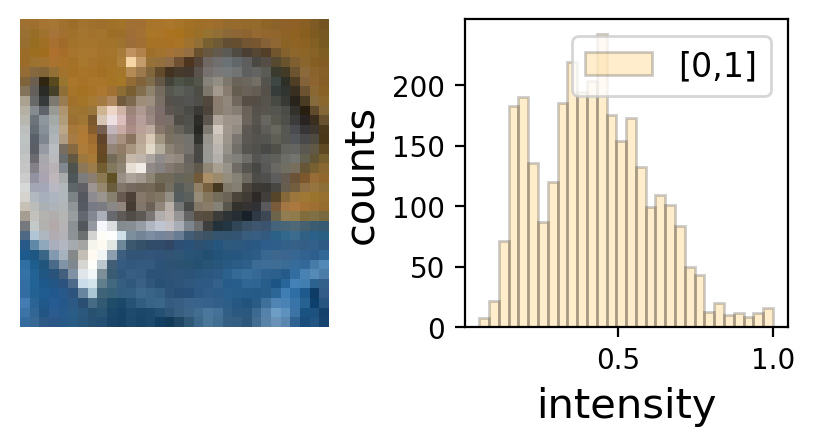}
    \caption{Example of CIFAR-10 data with centered and [0,1]-normalized image intensities.}
    \label{norm_effect_img}
\end{figure}

In our experiments, we noticed that data normalization had an effect on convergence stability of SGD in MLP models. We trained wide and deep MLP models on disjoint CIFAR-10 data subsets with equal amount of labels under a fixed SGD noise sample. Image intensities were either centered or normalized to the [0,1] range (\cref{norm_effect_img}). 

\subsubsection*{Discussion}
\begin{figure}[!ht]
    \centering
    \includegraphics[width=1\linewidth]{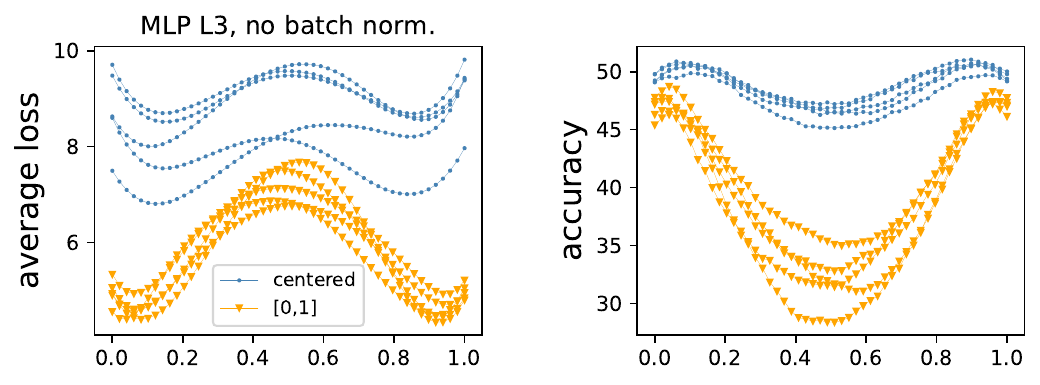}
    \caption{\textbf{Normalizing image intensities to the [0,1] range "breaks" LMC as opposed to data centering in MLPs without batch normalization.}\\
    Interpolation between MLP models with three hidden layers without batch normalization, trained on centered and [0,1]-normalized CIFAR-10 data subsets with equal amount of labels under the same training conditions (batch size 1024, learning rate $10^{-3}$, fixed SGD noise sample). Results from five experimental runs (five different seeds).}
    \label{normalization_interp}
\end{figure}

We observed that models, trained on [0,1]-normalized data subsets exhibited barrier and significant drop in accuracy as opposed to models, trained on centered data (\cref{normalization_interp}, Appendix \cref{tables}, table VIII).

\begin{figure}[!ht]
    \centering
    \includegraphics[width=1\linewidth]{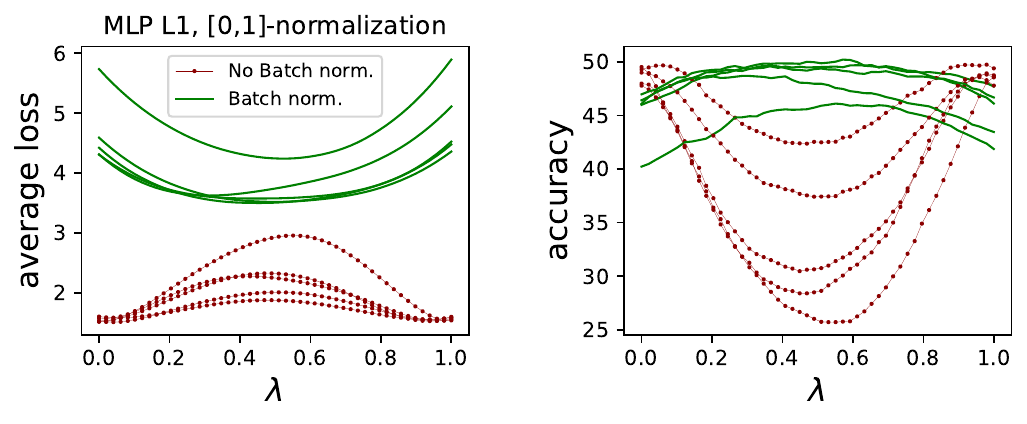}
    \caption{\textbf{Batch normalization layers "recover" LMC, even when models are trained on [0,1]-normalized data.}\\
    Interpolation between MLP models with one hidden layer followed by batch normalization layer, trained on [0,1]-normalized CIFAR-10 data subsets under the same training conditions (batch size 1024, learning rate $10^{-3}$, fixed SGD noise sample).}
    \label{normalization_bn_interp}
\end{figure}

MLP models with batch normalization layer(s), however, converged to the same local minimum, even when trained on [0,1]-normalized data (\cref{normalization_bn_interp}). We observed the same trend for ResNet and VGG models, which employ batch normalization layers.

We suggest that this effect arises because of the "partial loss of information" in terms of image features (\cref{norm_effect_img}). Since the weight distribution of initialized models include negative values, some information vanishes due to application of ReLU activation on the outputs of linear layers in case of [0,1]-normalized data. The loss of information may amplify covariate shift between training subsets, to the extent that alters optimization trajectory. This interpretation can also explain the learning curves, where one-hidden-layer MLPs learn the training data up to $60\%$ of accuracy in case of [0,1] normalization and up to $98\%$ in case of centered data (Appendix \cref{norm_lc_section}). The same trend is observed in deep MLPs, but less strongly. Adding batch normalization layer(s) restores stability in the case of [0,1] normalization, producing nearly identical learning curves. By standardizing mini-batch activations, followed by a learnable scale and shift, batch normalization aims to reduce internal covariate shift \emph{within} the network \cite{bn}. This stabilizes training (facilitating the emergence of LMC) and makes performance independent of the chosen data normalization scheme.

\subsection{Effects of Batch Size and Learning Rate on Models' Similarity}\label{section_bs_lr_effects}

We looked into similarities of the trained models and made the following observations regarding the effects of batch size and learning rate.

\begin{enumerate}
    \item \emph{Increasing batch size or decreasing learning rate results in more similar models.} The \cref{lr_bs_sim} shows scatter plots of angle, computed from cosine similarity measure, vs Manhattan distance. Metric values are comparable for models, trained in small batch \&  small learning rate and in large batch \& large learning rate regimes. 

    \begin{figure}[!ht]
        \centering
        \includegraphics[width=1\linewidth]{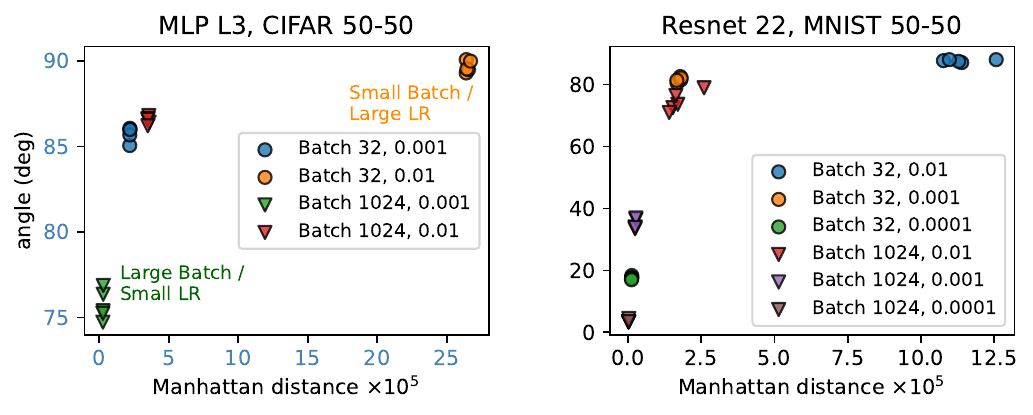}
        \caption{\textbf{In the large-batch and/or small-learning-rate regime, models are more similar, though not all exhibit LMC.}\\
        Angle between models' parameter vectors (from cosine similarity measure) vs Manhattan distance. Note that, MLP models, trained with learning rate 0.01, had batch normalization layer(s).}
        \label{lr_bs_sim}
    \end{figure}
    \item \emph{MLPs models can be can be nearly orthogonal and yet exhibit LMC} (\cref{angles}).
    
    \begin{figure}[!ht]
        \centering
        \includegraphics[width=0.6\linewidth]{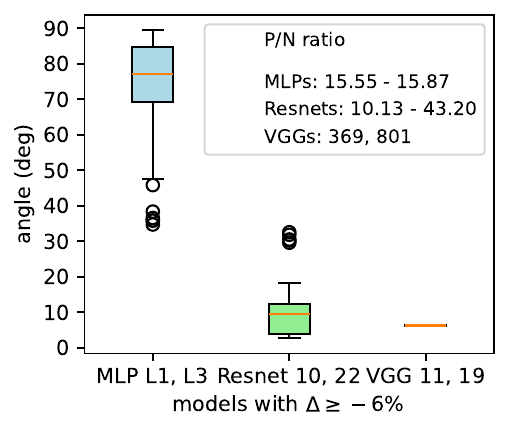}
        \caption{\textbf{MLP models can be nearly orthogonal and yet exhibit LMC.}\\Angle between a pair of models, that showed nearly-zero barrier and accuracy drop $\Delta \geq -6\%$ on the linear interpolation. Angle is computed from cosine similarity measure. Models have various depth, parameterization ($P/N$ ratio) and training conditions (batch size, learning rate, data).  Box plot is constructed from 127 (MLP), 80 (Resnet) and 15 (VGG) data points.}
        \label{angles}
    \end{figure}
    \item \emph{For some training conditions (batch size, learning rate) deeper models exhibit instability (loss barrier) within a comparable angular range as opposed to less deep models, exhibiting LMC.} We employ polar plots to simultaneously visualize instability (in terms of loss and accuracy) and the similarity of interpolated models (\cref{polar_diff_models}). Cosine similarity (angle) and Manhattan distance are measured between a pair of models, namely, $f_{\mathbf{\Theta}_A}$ and every interpolated model $f_{\mathbf{(1-\lambda) \mathbf{\Theta}_A + \lambda \mathbf{\Theta}_B}}$, $ \lambda \in \{0,...,1\}$. As can be seen, Manhattan distance and angle progressively increase with increasing $\lambda$ as it should. Note that, MLPs with one (L1) and three (L3) hidden layers have nearly the same parameterization regime ($P/N$ ratio, \cref{loss_landscape}, Appendix \cref{models}) and consequently, similar Manhattan distance.

    \begin{figure}[!ht]
        \centering
        \includegraphics[width=1\linewidth]{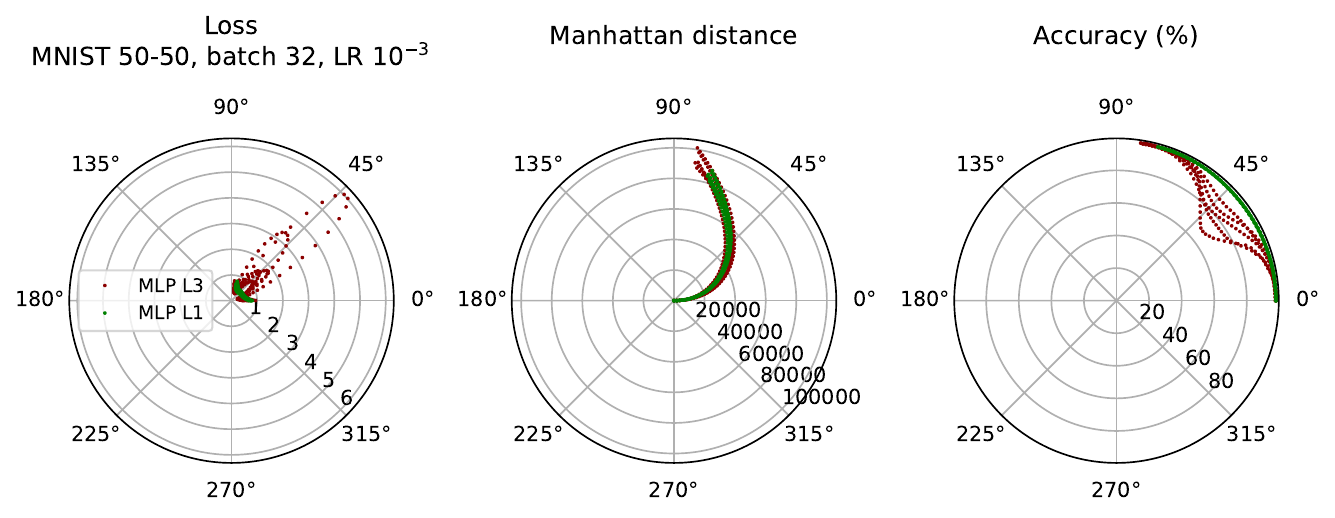}
        \includegraphics[width=1\linewidth]{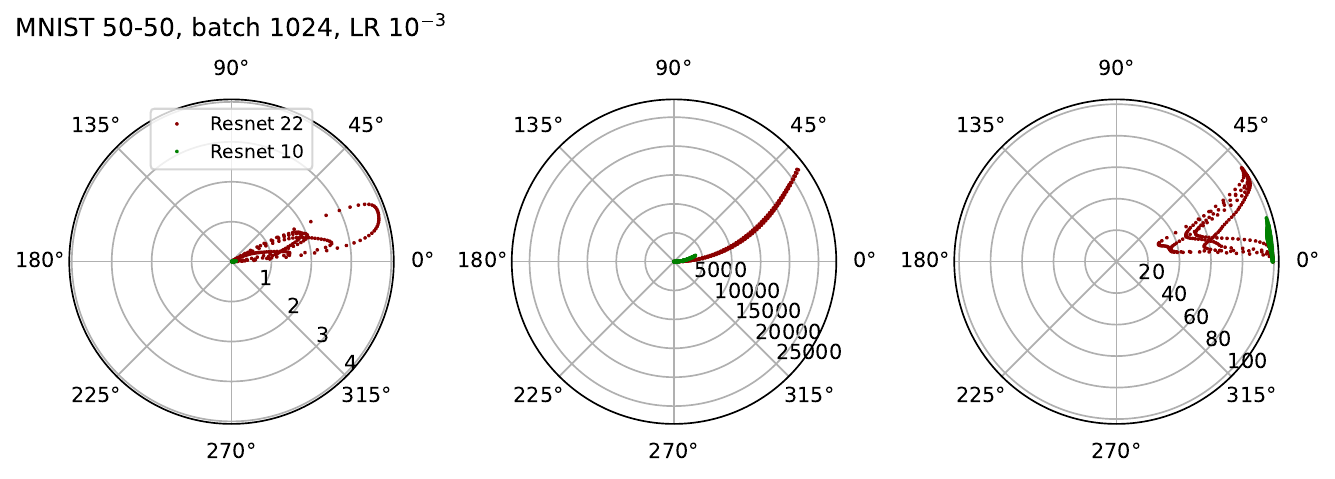}
        \includegraphics[width=1\linewidth]{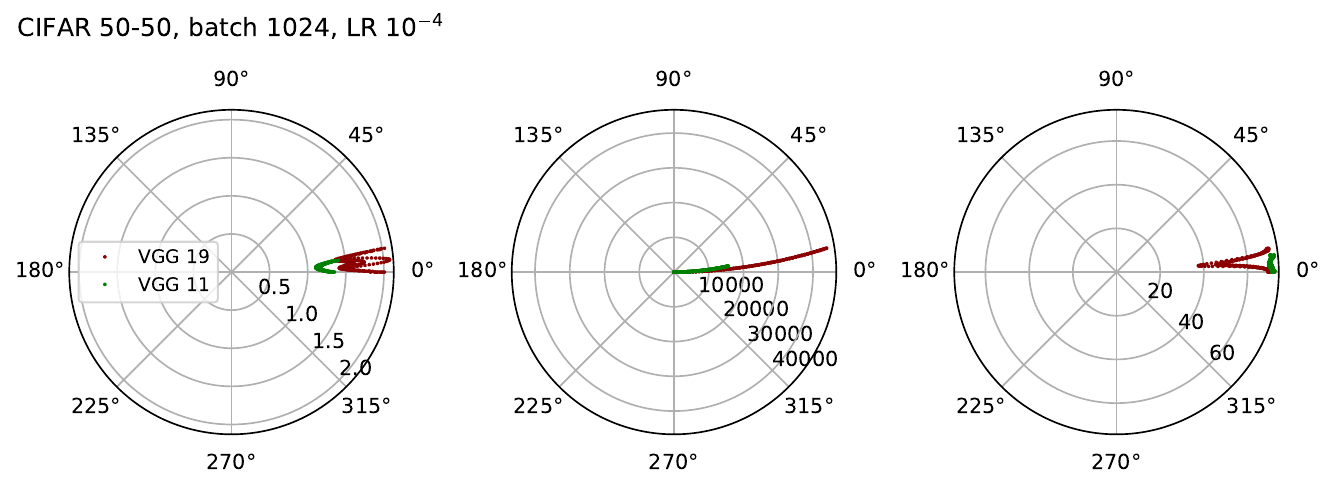}
        \caption{\textbf{Deeper models exhibit instability (loss barrier, accuracy drop) within a comparable angular range as opposed to less deep models, which exhibit LMC under the same training conditions.}\\
        Polar plot of loss, Manhattan distance and accuracy of interpolated models from five experimental runs for each architecture. ResNet and VGG of variable depth have different parameterization and, consequently, different Manhattan distance.}
        \label{polar_diff_models}
    \end{figure}
\end{enumerate}

\subsubsection*{\textbf{Discussion}}

Based on the experimental observations, the discussion is focused around the following questions:

\begin{itemize}
    \item \emph{Why increasing batch size and/or decreasing learning rate make models more similar (\cref{lr_bs_sim}) and "favors" LMC (\cref{batch_lr_tradeoff})?} 
    \item \emph{If instability is present, why does value of barriers depends on batch size and learning rate (\cref{batch_lr_tradeoff})?} 
    \item \emph{In case of small values of barriers (0.1 to 1) but accuracy drop up to 15$\%$, do linear interpolation plots actually indicate training stability? What explains such a drop in performance (\cref{interpolations_bigearth}, Appendix \cref{tables}, table VI)?}
    \item \emph{Why some models, exhibiting LMC, show "narrow angular range" while MLPs can be nearly orthogonal (\cref{polar_diff_models})?} 
\end{itemize}

\subsubsection*{\textbf{Training stability due to vanishing SGD noise}}
As is well established, mini-batch training yields a noisy estimate of the true loss gradient and, consequently, noisy updates of the model parameters. Smith et al. proposed that scale of noise $g$ due to mini-batch training depends on batch size $B$, learning rate $\varepsilon$ and size of training data set $N$ \cite{smith}: 

\begin{equation*}
    g = \varepsilon (\frac{N}{B}-1)
\end{equation*}
which means, in large-batch or small-learning rate regime noise vanishes ($\lim_{B \rightarrow N} g = 0$ and $\lim_{\varepsilon \rightarrow 0} g = 0$). In some of our experiments we trained two copies of the same model on disjoint subsets \emph{with the same data order} (the same labels / image semantics, \cref{scheme}), i.e. fixed sample of SGD noise \cite{frankle}, and observed instability in \emph{small}-batch, \emph{large}-learning rate regime (\cref{interp_50_50}). \emph{The results suggest that gradient estimates should have high variance due to covariate shift, which can be seen as a different origin of SGD noise.} Increasing batch size or decreasing learning rate reduces noise from data shifts, varying data orders and augmentation, leading to more stable training. In such cases, models follow similar trajectories through the loss landscape and converge to the same local minimum or region, resulting in greater similarity (smaller angles and smaller Manhattan distances).

\subsubsection*{\textbf{Dependence of barriers on learning rate and batch size: "nature" of local minima}}

\begin{figure}[!ht]
    \centering
    \includegraphics[width=0.45\linewidth]{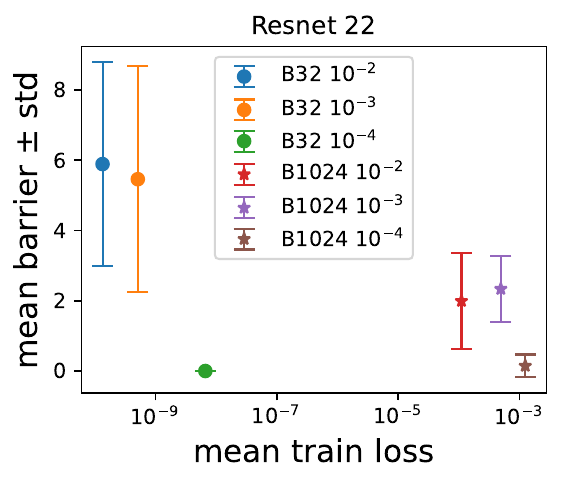}
    \includegraphics[width=0.42\linewidth]{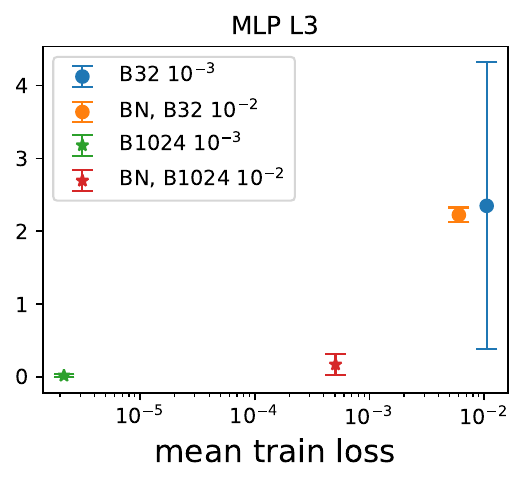}
    \caption{\textbf{The barrier decreases as batch size increases or learning rate decreases. Then the resulting variation in training loss values suggests convergence to different regions of loss landscape, characterized by different degrees of smoothness.}\\
    Mean and standard deviation are obtained from five experimental runs (\emph{the same} set of seeds). Training loss is also averaged over values from two training subsets. Notation: BN stands for batch normalization layer.}
    \label{barrier_trainloss}
\end{figure}

Entezari et al. shown experimentally that barrier increases with \emph{depth} \cite{entezari}. This is thought to be related with increasing "non-convexity" or "chaotic" behavior of loss landscape due to over-parameterization. We find that, in addition, barriers are influenced by learning rate and batch size (also in \cite{alt}): increasing batch size or decreasing learning rate reduces barrier and associated drop in performance of interpolated models. \emph{This indicates that models converge to regions of loss landscape characterized by different degrees of smoothness.} The \cref{barrier_trainloss} clearly shows different loss values under different training conditions.\\
For ResNet and VGG models, larger batch sizes and smaller learning rates lead to higher final (training) loss values. The opposite trend in case of batch size is observed for MLP models.  

Furthermore, since large-batch training affects SGD convergence towards \emph{sharp minima} \cite{visualizing}, which are correlated with poorer generalization, this could explain (1) small barrier values (or small loss fluctuations) but accuracy drops up to $15\%$ and (2) small angular range (high cosine similarity) of models \emph{in case of large-batch training}. MLP models, however, can become nearly orthogonal in large-batch training, possibly due to the influence of model architecture on the loss landscape.

\subsection{Pros and cons of Linearly Mode Connected Deep Ensemble}

\begin{table*}[!ht]
    \begin{tabular}{|c|cc|cc|cc|}
    \hline\hline
    Model & Resnet 22 & & VGG 11 & & MLP L1 & \\
    CIFAR 50-50 & LMC$^*$ & Diff. seeds$^{**}$ & LMC & Diff. seeds & LMC & Diff. seeds \\\hline\hline
    $\overline{WA}$  & $\mathbf{0.197\pm0.004}$ & $0.105$ & $\mathbf{0.196\pm0.003}$ & $0.112$ & $0.245\pm0.002$ & $0.185$\\[1.5ex]\hline
    $\overline{WD}$ & $0.045\pm0.003$ & $0.081$ & $0.043\pm0.002$ & $0.076$ & $0.165\pm0.002$ & $0.206$\\[1.5ex]\hline
    \begin{tabular}{c}
        Majority vote\\
        test acc. (\%)
    \end{tabular}& $71.16\pm0.73$ & $\mathbf{77.46}$ & $71.63\pm0.43$ & $\mathbf{77.39}$ & $50.39\pm0.41$ & $52.13$ \\[1.5ex]\hline
        \begin{tabular}{c}
        Averaged predictions\\
        test acc. (\%)
    \end{tabular}& $72.18\pm0.7$ & $\mathbf{78.16}$ & $72.94\pm0.34$ & $\mathbf{78.02}$ & $52.43\pm0.35$ & $53.8$ \\\hline\hline
    \end{tabular}
    \caption{\textbf{Resnet and VGG models from the same local minimum (exhibiting LMC) make the same mistakes more consistently than models from different initializations. Consequently, ensemble of the latter has higher test accuracy. This trend is less pronounced in MLPs, potentially on account of the low similarity of models, albeit from the same local minimum.}\\
    $^*$Results are averaged over five to ten ensembles.\\
    $^{**}$Results are averaged over 2 ensembles.}
    \label{table_pros_cons}
\end{table*}

\begin{figure*}[!ht]
    \centering
    \includegraphics[width=0.2\linewidth]{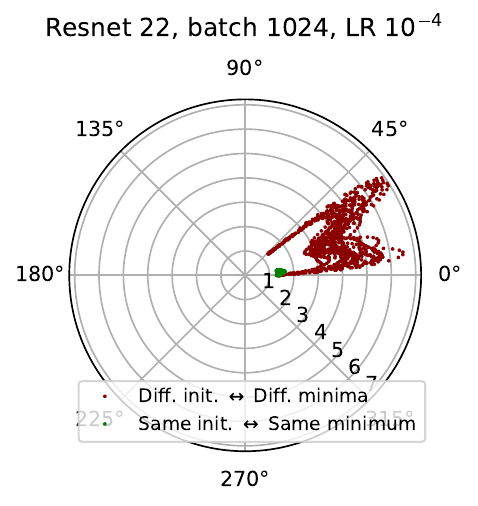}
    \includegraphics[width=0.2\linewidth]{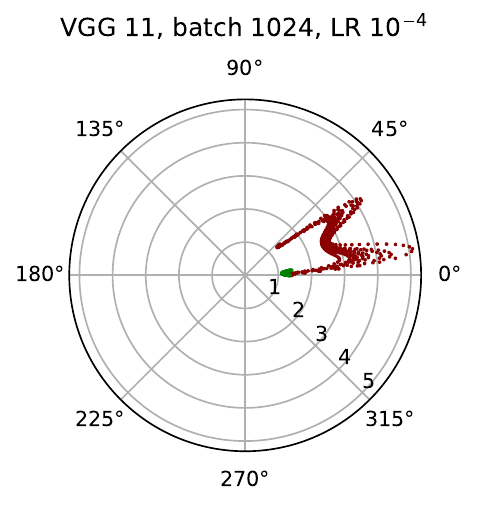}
    \includegraphics[width=0.2\linewidth]{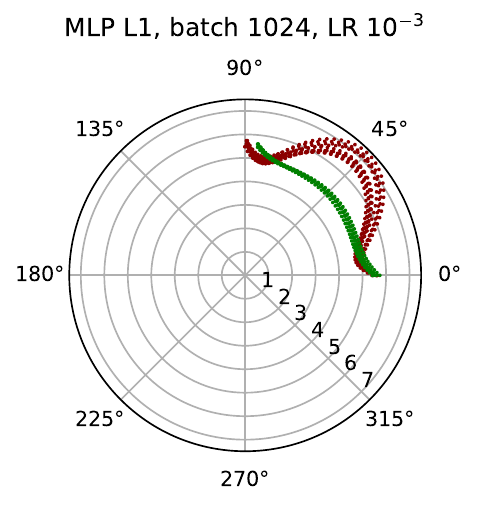}
    \includegraphics[width=0.3\linewidth]{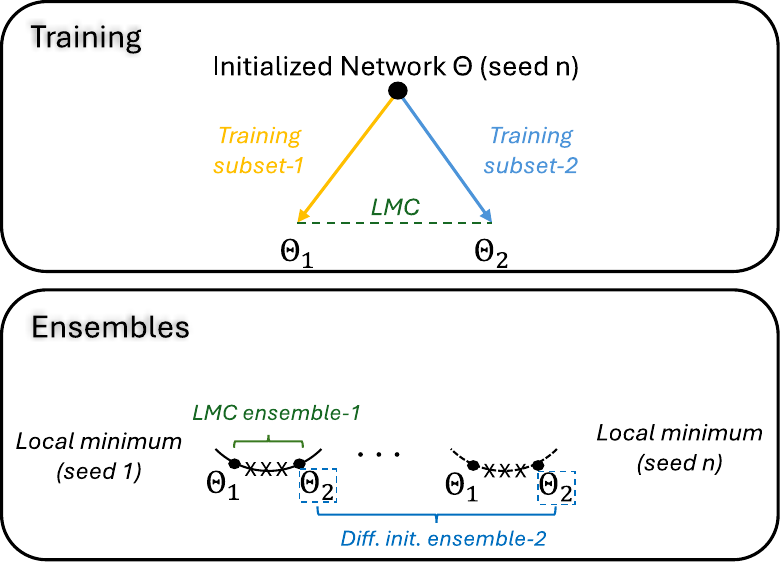}
    \caption{Polar plot of loss of interpolated models from five to ten experimental runs (seeds). \textbf{Models, trained from the same initialization on disjoint subsets exhibit LMC, so linear interpolation shows nearly constant loss (and accuracy). Models, trained from different initializations on the same subset, show loss barrier(s) on linear interpolation, which confirms they are in different local minima}. \textbf{MLPs are nearly orthogonal in both cases.}
    Schematic illustration of training procedure and ensembles construction.}
    \label{same_diff_init}
\end{figure*}

The \cref{table_pros_cons} reports average wrong agreement $\overline{WA}$, average wrong disagreement $\overline{WD}$ and test accuracy of ensembles of models, sampled "via LMC" and trained from different initializations. 

\subsubsection*{Details}
Each model architecture (Resnet 22, VGG 11, MLP L1) was trained on two disjoint subsets of CIFAR 50-50 (\cref{partitions}) five to ten times (different seeds). Ensembles were constructed from interpolated models and models, trained from different initializations as schematically illustrated in the \cref{same_diff_init}. We evaluate test accuracy of ensembles from predictions, obtained by majority vote and by averaging models' prediction vectors.

\subsubsection*{Discussion}

Average wrong agreement for Resnet, VGG models, sampled via LMC, is about twice as large as for models, trained from different initializations (\cref{table_pros_cons}). Naturally, average wrong disagreement of the latter is larger. \emph{This means that models from the same local minimum make the same mistakes more consistently than models from different local minima, which agrees with the conclusion of \cite{fort} about functional diversity of the latter.} Consequently, ensemble of models, trained from different initializations, has higher test accuracy, as evaluated both from majority vote and averaged predictions. These observations are, however, less pronounced in MLPs. Possible reason for it could be overfitting or because MLP models in general are nearly orthogonal and so significantly less similar as can be seen from the polar plot of loss of interpolated models (\cref{same_diff_init}, accuracy and Manhattan distance are presented in the Appendix, \cref{polar_plots}). 

\section{Conclusion}
We experimentally investigated the effects of data shifts (covariate shift, label imbalance, domain shift) on stability of mini-batch training dynamics of models of different architecture (MLP with and without batch normalization, ResNet and VGG). These effects can be mitigated by the size of a training mini-batch and a learning rate, affecting models' similarities and convergence to either to the same local minimum or regions of different smoothness and generalization properties. We observed that models sampled via LMC tend to make similar errors more frequently than models converged to different local minima. Since ensemble performance grows with its size, the benefit lies within a trade-off between training many models versus two, exhibiting LMC.\\
We summarize the following known and speculative relations, with the latter indicating directions for future research:

\begin{enumerate}
    \item \textbf{Training stability - SGD noise scale relation:} training stability is affected by SGD noise scale due to mini-batch training \cite{smith}. Training remains stable when the learning rate, $\varepsilon$ or the data size-to-batch ratio, $\frac{N}{B}$ is small, leading to LMC.
    \item \textbf{Batch-size - sharpness - generalization relation: }large-batch and small-batch training influences convergence towards sharp and flat local minima, respectively, sharper minima being correlated with poorer generalization \cite{visualizing}.
    \item \emph{\textbf{Angular range $ \stackrel{?}{-}$ sharpness / basin volume relation}: is  it legitimate to relate angular range (cosine similarity) of models, exhibiting LMC, to basin sharpness or its volume (definitions are discussed in \cite{dinh})? That is, is narrow angular range a specificity of optimization, model architecture or basin volume?} 
    \item \emph{\textbf{Sharpness / basin volume $ \stackrel{?}{-}$ functional diversity relation; robustness}: if SGD solutions are sampled from a sharp minimum, does it imply reduced or greater functional diversity as compared to a flat minimum? Are models more robust?} 
\end{enumerate}

\section*{Acknowledgments}
We thank J. Gawlikowski, M. Ortiz Torres and B. Franke for fruitful discussions and support of this work.

\section*{Statement}
ChatGPT by OpenAI was used for language editing and grammar enhancement to improve readability.


\clearpage
\appendices

\section{BigEarthNet Bands}\label{bigearth_data}
BigEarthNet Sentinel-2 bands or their combinations, used in this work, are indicated in the following table in \textbf{bold}. VNIR and SWIR stand for visible \& near infrared and short wave infrared, respectively.

\begin{table}[!ht]
    \resizebox{\linewidth}{!}{
    \begin{tabular}{|c|l|c|c|}
    \hline\hline
    Band & \begin{tabular}{c}
        Resolution \\
        (matrix size) 
    \end{tabular} & \begin{tabular}{c}
        Central\\
        Wavelength\\
        (nm) 
    \end{tabular} & Description\\\hline\hline
    $B_1$ & 60 m (20x20) & 443 & ultra blue \\\hline
    $\mathbf{B_2}$ & 10 m (120x120) & 490 & \textbf{blue} \\\hline
    $\mathbf{B_3}$ & 10 m (120x120) & 560 & \textbf{green}\\\hline
    $B_4$ & 10 m (120x120)	& 665 & Red\\\hline
    $B_5$ & 20 m (60x60) & 705 & VNIR\\\hline
    $B_6$ & 20 m (60x60) & 740 & VNIR\\\hline
    $B_7$ & 20 m (60x60) & 783	& VNIR\\\hline
    $\mathbf{B_8}$ & 10 m (120x120) & 842 & VNIR\\\hline
    $B_{8a}$ & 20 m (60x60) & 865 & VNIR\\\hline
    $B_9$ & 60 m (20x20) & 940 & SWIR\\\hline
    $B_{10}$ & 60 m & 1375 & SWIR\\\hline
    $B_{11}$ & 20 m (60x60) & 1610 & SWIR\\\hline
    $B_{12}$ & 20 m (60x60) & 2190 & SWIR\\\hline
    $\mathbf{B4, B3, B2}$ & & & \textbf{Natural color}\\\hline
    $\mathbf{B8, B4, B3}$ & & & \textbf{Color infrared}\\\hline
    $\mathbf{B11, B8, B2}$ & 10 m (120x120)*	 & & \textbf{Agriculture}\\\hline
    \hline\hline
    \end{tabular}}
    \caption{\textbf{Description of bands of BigEarthNet dataset.}\\*B11 images were up-sampled.}
\end{table}

\section{Fixing sample of SGD noise}\label{pseudocode}

When training data are split into subsets with an equl number of labels, we ensure identical data ordering across subsets to fix the SGD noise sample during training. Introducing label imbalance, however, alters the data order in both subsets, resulting in different SGD noise samples. The pseudo-code below illustrates how the SGD noise sample is fixed during training with subsets. Since any random operation (model initialization, data shuffling, horizontal image flip, dropout) changes the state of pseudo-random number generator (RNG), subsequent calls produce different random numbers. Resetting the state of RNG guarantees the same SGD noise sample. In the implementation, this is done in two ways to allow for the following flexibility:

\begin{algorithm}
\caption{Training on data subsets with fixed sample of SGD noise}
\begin{algorithmic}[1]
\begin{small}
\Function{SET SEED}{seed}        
    \State for pytorch RNG for all devices $\&$ in other libraries
\EndFunction
\Statex
\State Create training, test datasets
\State Subset 1, Subset 2 $\gets$ create subsets(training dataset, split $\%$)
\State Define Train dataloader 1 (Subset 1)
\State Define Train dataloader 2 (Subset 2)
\State Define Test dataloader (test dataset)
\Statex
\State SET SEEDS(seed) \Comment{\emph{for models' initialization}}
\State Initialize Model 1
\State Model 2 $\gets$ deepcopy(Model 1)
\State Define Optimizer 1, Optimizer 2
\Statex
\Procedure{Training on Subset 1}{}

\State SET SEEDS(seed)
    \For{epoch$= 0$ to $n$}
        \If{SGD noise == 'same'} \Comment{\emph{comment if necessary}}
            \State torch.manual seed(epoch)
        \EndIf
        \State train(Model 1, Train loader 1, Optimizer 1), track labels
        \State test(Model 1, Test loader)
    \EndFor
\EndProcedure
\Statex
\Procedure{Training on Subset 2}{}
    \If{SGD noise == 'same'}
        \State SET SEEDS(seed)
    \Else
        \State new seed = seed + 100
        \State SET SEEDS(new seed)
    \EndIf
    \For{epoch$= 0$ to $n$}
        \If{SGD noise == 'same'} \Comment{\emph{comment if necessary}}
            \State torch.manual seed(epoch)
        \EndIf
        \State train(Model 2, Train loader 2, Optimizer 2), track labels
        \State test(Model 2, Test loader)
    \EndFor
\EndProcedure
\end{small}
\end{algorithmic}
\end{algorithm}

\begin{enumerate}
    \item RNG is reset within epoch loop by torch.manual seed(epoch). This guarantees to have the same random operations not only during training with subset 1 and subset 2, but also across different experimental runs (seeds).
    \item Commenting the if-statement in epoch loop will yield different SGD noise samples across different experimental runs (seeds), but the same SGD noise sample across the two train loaders, when specified.
\end{enumerate}

In case of small datasets, we load data in the main process\footnote{\href{https://docs.pytorch.org/docs/stable/notes/randomness.html}{https://docs.pytorch.org/docs/stable/notes/randomness.html}}, and we use multiple workers with fixed seed for BigEarthNet. We set torch.backends.cudnn.benchmark = True to improve performance, although CUDA convolution operations can introduce nondeterminism across executions. 

\subsection*{Checks}

\begin{figure}[!ht]
    \centering
    \includegraphics[width=1\linewidth]{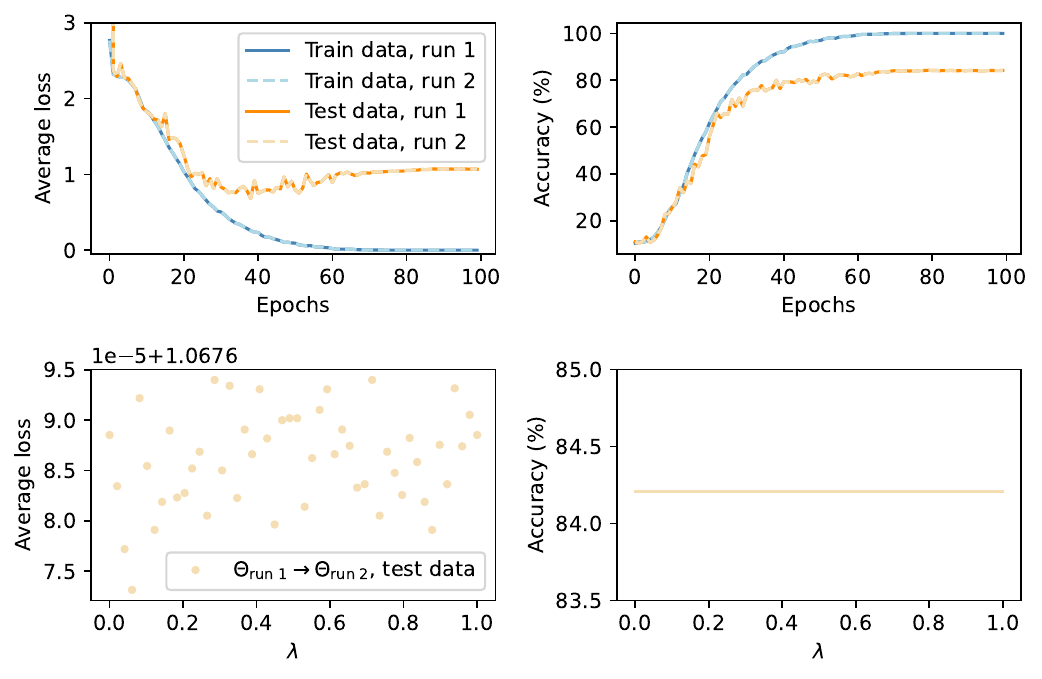}
    \caption{\textbf{Reproducibility of two identical experimental runs.}\\
    Average losses and accuracy (\textbf{top}). Linear interpolation between models from two identical runs (\textbf{bottom}): average loss and accuracy (evaluation on test data).}
    \label{vgg19_checks}
\end{figure}

\begin{itemize}
    \item During training, we track the first 10 labels of batch number 10 in epochs 1 and 20 for each training dataloader. In case of fixed SGD noise sample and data split 50/50 (\cref{partitions}) the labels are identical across the train dataloaders. Files generated during the experiments are provided in the supplementary materials.
    \item We perform the same experiment twice (the same seed) and compare models trained with the same subset: the results are reproducible (loss, accuracy values) and linear interpolation yields effectively constant accuracy and loss values with fluctuations on the order of $10^{-5}$ or $ 10^{-8}$, which can be attributed to minor numerical variations inherent in CUDA computations (\cref{vgg19_checks}).
\end{itemize}

\section{Models}\label{models}
The model architectures, total number of parameters ($P$) and parameterization ratios are reported in the table V. The training subsets contain 29,997 samples for MNIST and 25,000 for CIFAR-10. During training CIFAR data undergo random augmentation (random horizontal image flip). Each model was trained (mostly) five times with different initialization seeds: 1, 20, 100, 500, and 1500. MLP models with one (L1) and three (L3) hidden layers were trained both with and without batch normalization (BN). Parameterizations of the latter are not fully listed, as their ratios are similar.\\
We used the Adam optimizer with negative log-likelihood loss for training on MNIST and CIFAR-10, and binary cross-entropy with logits loss for BigEarthNet.

\begin{table}[!ht]
    \resizebox{\linewidth}{!}{
    \begin{tabular}{|l|c|c|c|}
    \hline\hline
    Model & Dataset & $P$ & $P/N$ ratio\\\hline\hline
    MLP L1 W587 & MNIST & 466 675 & 15.56\\
    MLP L1 W587, BN & & 467 849 & 15.60\\
    MLP L3 W256-512-256 & & 466 442 & 15.55\\
    MLP L1 W126 & CIFAR-10 & 388 468 & 15.54\\
    MLP L3 W110-256-110 & & 395 826 & 15.83\\\hline
    ResNet10 & MNIST & 303 866 & 10.13 \\
    ResNet22 & & 1 079 802 & 36.00 \\
    ResNet10 & CIFAR-10 & 304 154 & 12.16 \\
    ResNet22 & & 1 080 090 & 43.20\\\hline
    VGG11 & CIFAR-10 & 9 231 114 & 369.24 \\
    VGG19 & & 20 040 522 & 801.62 \\\hline
    VGG-like & BigEarthNet & 1 082 771 & 2.82 \\
    \hline\hline
    \end{tabular}}
    \label{parameterizations}
    \caption{\textbf{Models and their parameterization ratios.} W indicates number of neurons in each hidden layer of an MLP model.}
\end{table}

\newpage
\section{Definitions of similarity metrics}\label{sim_metrics}

\subsection*{Cosine similarity} 

Cosine similarity assesses directional alignment of two models' parameter vectors $\mathbf{\Theta}_A$ and $\mathbf{\Theta}_B$, expressed in matrix form for each layer as

\begin{equation*}
    \cos \angle (\mathbf{\Theta}_A, \mathbf{\Theta}_B) = \frac{1}{N_A N_B}\sum^L_{l=1} \Big\langle \mathbf{W}^A_l,\mathbf{W}^B_l \Big\rangle_F
\end{equation*}
Notation: $l=1,...,L$ indicates a hidden layer, $\mathbf{W}_l$ represents  parameter matrix of the $l$th layer, including bias, $\langle \cdot, \cdot \rangle_F$ is Frobenius product of two real-valued matrices, $\langle \mathbf{A}, \mathbf{B} \rangle_F = \sum_{ij} a_{ij}b_{ij}$ (with $\langle \mathbf{A}, \mathbf{A} \rangle_F = \norm{\mathbf{A}}^2_F$ Frobenius norm of $\mathbf{A}$ \cite{strang}). $N$ is the norm of model parameter vector,\\ $N^2=\norm{\mathbf{\Theta}}^2_F = \sum^L_{l=1} \Big\langle \mathbf{W}_l, \mathbf{W}_l \Big\rangle_F = \sum_{lij} (w^l_{ij})^2$. 

\subsection*{Manhattan or Euclidean distance} 

In high-dimensional space the notion of proximity to a point becomes less meaningful as there is a poor discrimination between nearest and farthest neighbor to a given query point \cite{aggarwal}. The relative contrast of distances to a query point depends on the metric used and is better for Manhattan than for the Euclidean distance.  Manhattan distance ($k=1$) or square of Euclidean distance ($k=2$) between two models' parameter vectors $\mathbf{\Theta}_A$ and $\mathbf{\Theta}_B$, can be expressed as:

\begin{equation*}
    D^k_{AB} = \norm{\mathbf{\Theta}_A - \mathbf{\Theta}_B}^k= \sum_{lij} \vert w^{A,l}_{ij} - w^{B,l}_{ij}\vert^k
\end{equation*}

\section{Average wrong agreement and disagreement}\label{wa_wd}

Given $M$ models in an ensemble, we compare predictions of \emph{every pair} of models and the corresponding ground truth for all $N$ test data samples:

\begin{itemize}
    \item \emph{Average Wrong Agreement} is the number of times, when every two models agree in their predictions and both are wrong, normalized by the total count of pair-wise comparisons:

    \begin{equation*}
    \begin{array}{ll}
            \overline{WA} = & \frac{1}{C^M_2 N} \sum^M_{i<j} \sum^N_{n=1} \Big[f(x_n; \mathbf{\Theta}_i) = f(x_n; \mathbf{\Theta}_j)\Big] \cdot \\[3ex]
            & \frac{1}{2}\Big(\Big[f(x_n; \mathbf{\Theta}_i) \neq y_n\Big] + \Big[f(x_n; \mathbf{\Theta}_j) \neq y_n\Big]\Big)
    \end{array}
    \end{equation*}
    where $f(x_n; \mathbf{\Theta}_j)$ denotes prediction of model $\mathbf{\Theta}_j$, given a data sample (input) $x_n$, $y_n$ is ground truth label of $x_n$, $C^M_2=\frac{M!}{2!(M-2)!}$ is the number of combinations of every two models. Each equality or inequality in the brackets counts as 1.
    \item \emph{Average Wrong Disagreement} is the number of times, when every two models disagree in their predictions and both are wrong, normalized by the total count of pair-wise comparisons:
    \begin{equation*}
    \begin{array}{ll}
            \overline{WD} = & \frac{1}{C^M_2 N} \sum^M_{i<j} \sum^N_{n=1} \Big[f(x_n; \mathbf{\Theta}_i) \neq f(x_n; \mathbf{\Theta}_j)\Big] \cdot \\[3ex]
            & \frac{1}{2}\Big(\Big[f(x_n; \mathbf{\Theta}_i) \neq y_n\Big] + \Big[f(x_n; \mathbf{\Theta}_j) \neq y_n\Big]\Big)
    \end{array}
    \end{equation*}
\end{itemize}

For illustration the following matrices represent predictions of three data samples by five models, $P$ and the ground truth labels, $G$:

\begin{equation*}
    P = \overbrace{\begin{pmatrix}
        2&2&2&2&2\\
        6&6&7&8&8\\
        5&5&1&12&3
    \end{pmatrix}}^{M_1 ~ M_2 ~ M_3 ~ M_4 ~ M_5},~
    G = \begin{pmatrix}
        1\\7\\5
    \end{pmatrix}
\end{equation*}
There are 12 times, where every two models agree and are wrong; 7 times, where every two models disagree and are wrong; $N \cdot C^M_2 = 3 \cdot C^5_2=30$ the total number of element-wise comparisons made, which results in average wrong agreement of $0.4$ and average wrong disagreement of $0.23(3)$. In case when all models agree in their predictions and are wrong, $\overline{WA}=1$ and when all models predict every sample correctly, $\overline{WA}=\overline{WD}=0$.

\section{Label Imbalance effect}

\newpage
\section{CIFAR-10 Normalization}\label{norm_lc_section}

\begin{figure}[!ht]
    \centering
    \includegraphics[width=1\linewidth]{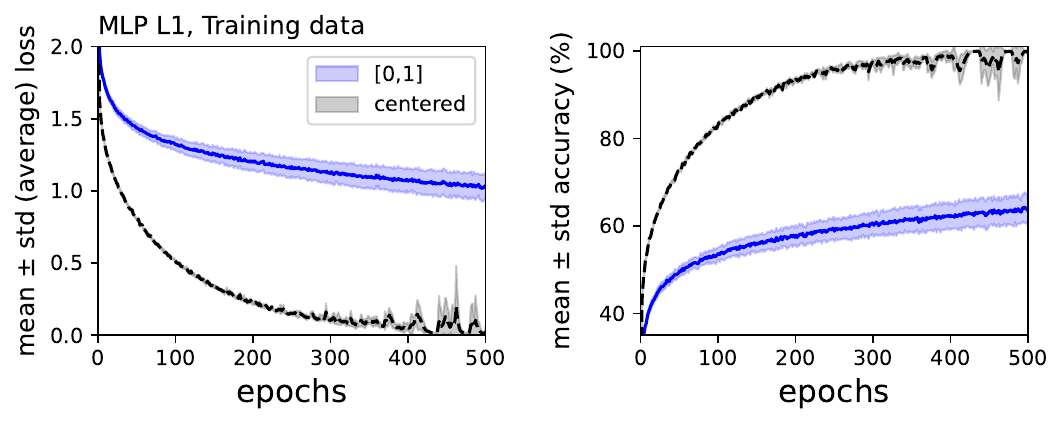}
    \includegraphics[width=1\linewidth]{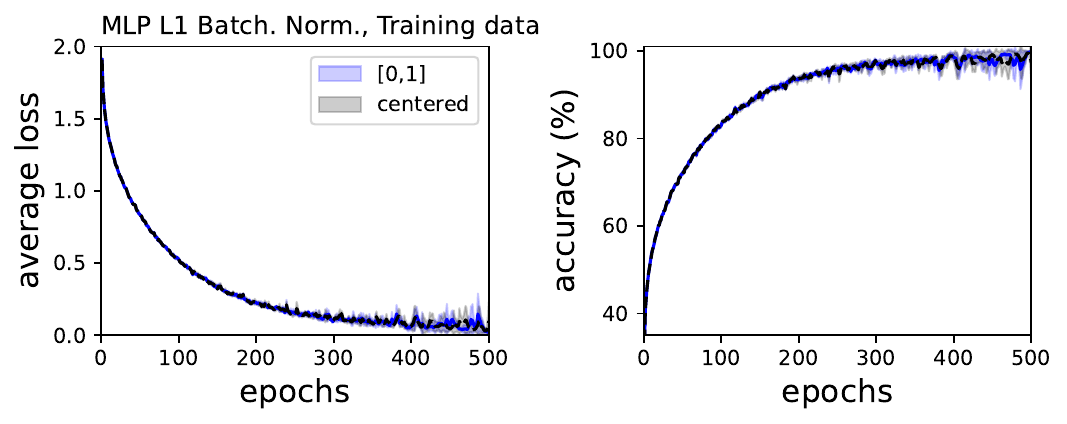}
    \caption{\textbf{Adding batch normalization layer leads to practically identical learning curves in case of centered and [0,1]-normalized data.}\\
    Learning curves for one-hidden-layer MLP without (\textbf{top}) and with batch normalization (\textbf{bottom}), trained on centered and [0,1]-normalized data (five experimental runs; mean curves with standard deviation (shaded)). Average loss and accuracy reported for one \emph{training} data subset. Models were trained with batch size 1024 and learning rate of $10^{-3}$.}
    \label{norm_effect_lc}
\end{figure}

\begin{figure}[!ht]
    \centering
    \includegraphics[width=1\linewidth]{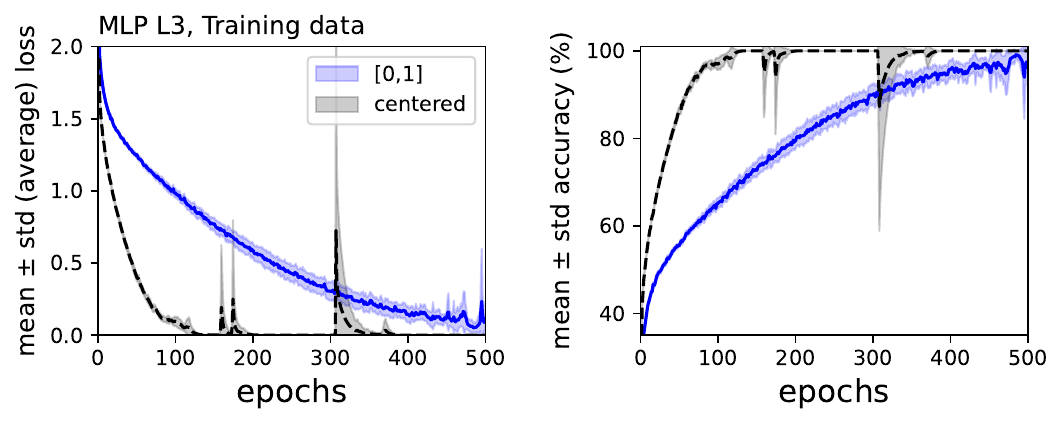}
    \includegraphics[width=1\linewidth]{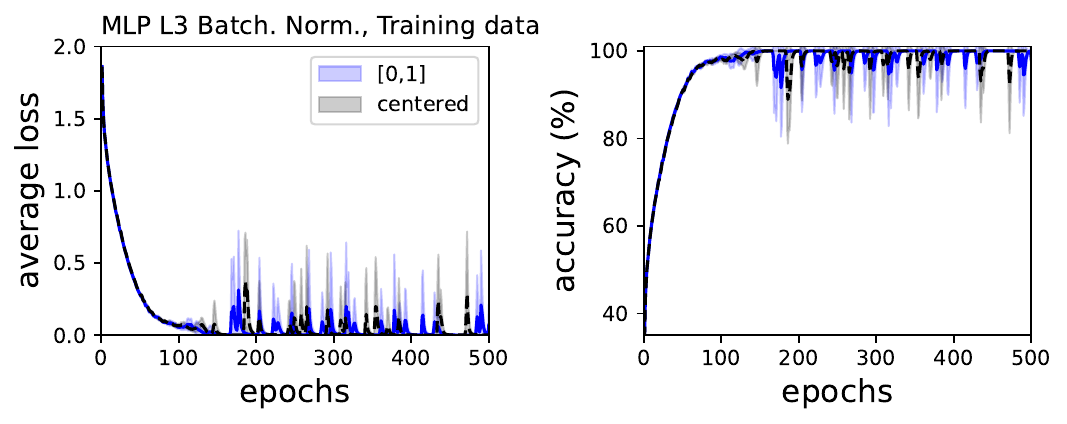}
    \caption{Same as in the \cref{norm_effect_lc}, but for three-hidden-layer MLPs.}
\end{figure}

\newpage
\section{Polar plots}\label{polar_plots}

    \begin{figure}[!ht]
        \centering
        \includegraphics[width=1\linewidth]{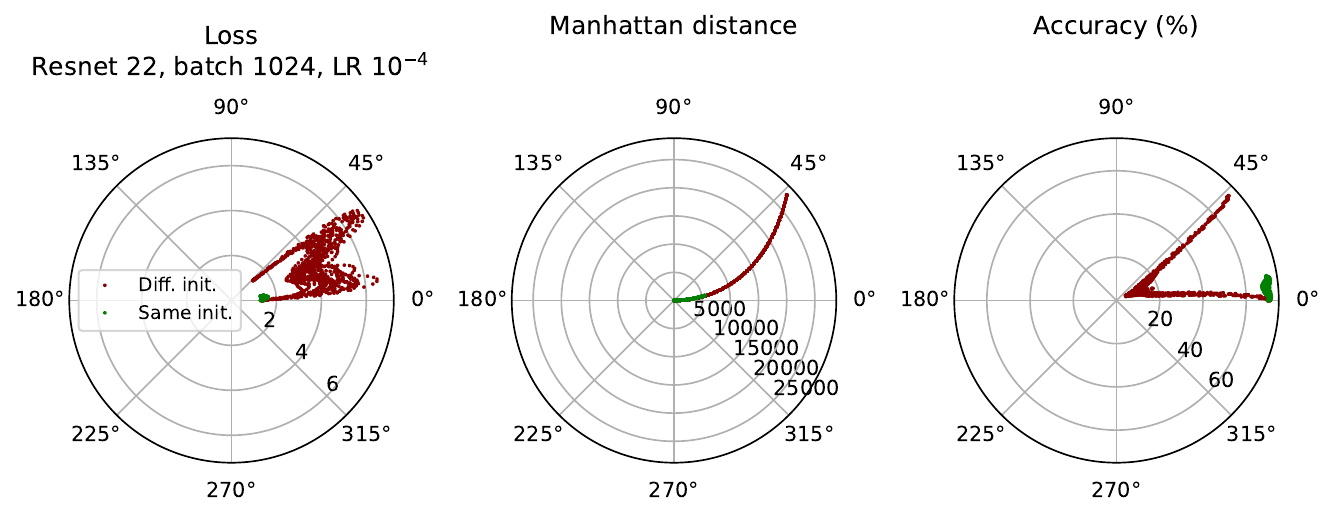}
        \includegraphics[width=1\linewidth]{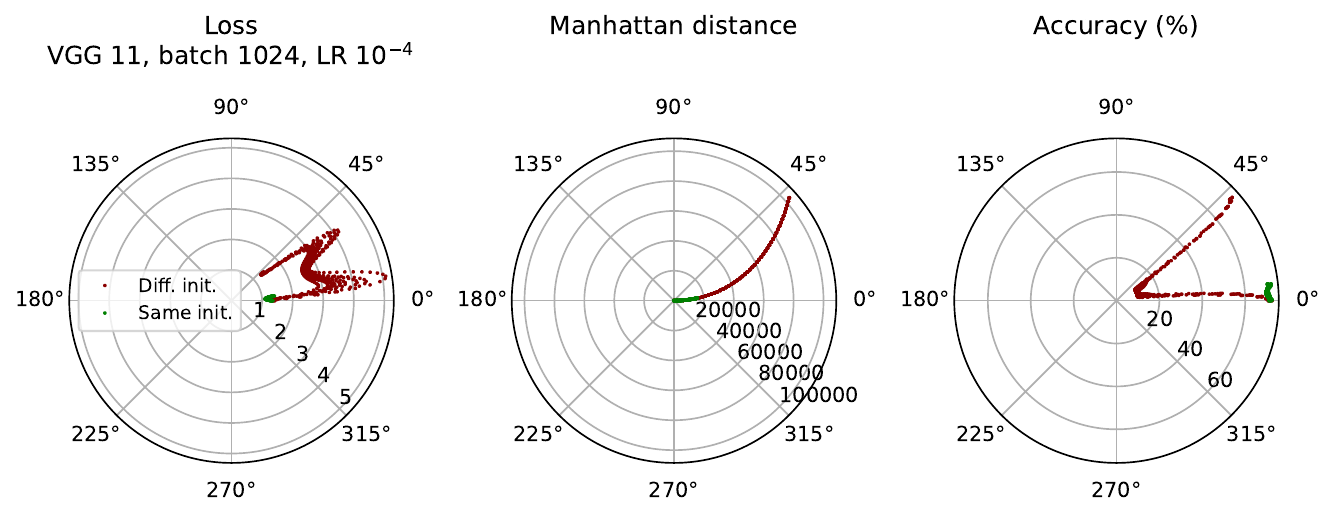}
        \includegraphics[width=1\linewidth]{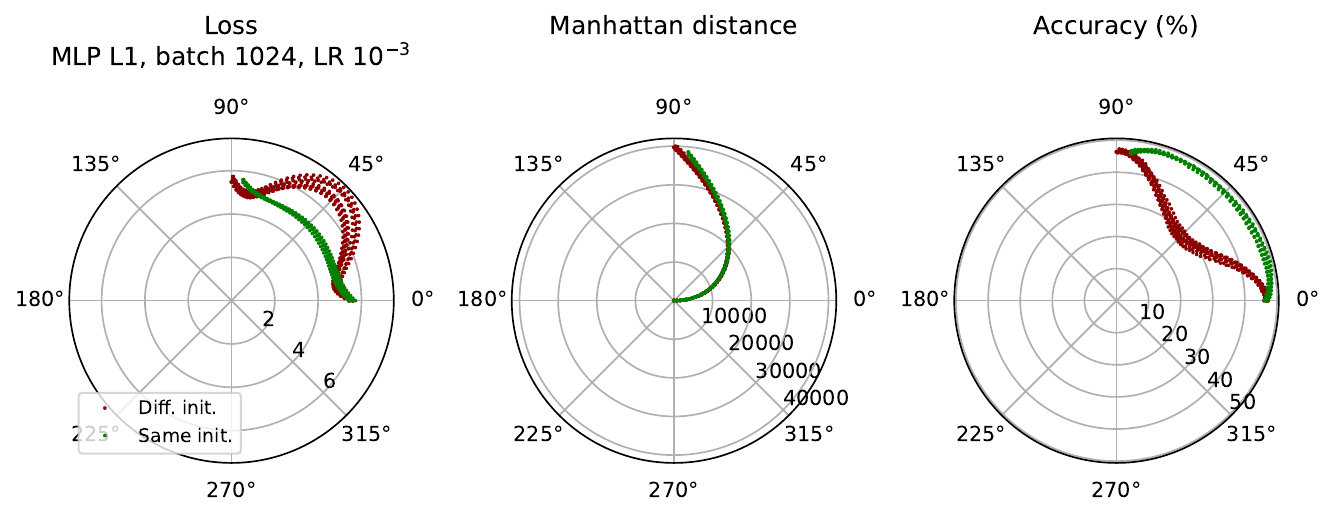}
        \caption{Polar plot of loss, Manhattan distance and accuracy of interpolated models from five to ten experimental runs (seeds).}
    \end{figure}


\section{Tables}\label{tables}

\begin{table*}[!ht]
    \resizebox{\linewidth}{!}{
    \begin{tabular}{|l|c|c|ll|ll|}
    \hline\hline
    Model & Data & LR & Barrier & & $\Delta~(\%)$ &\\
    & & & Batch 32 & Batch 1024 & Batch 32 & Batch 1024\\[1.5ex]
    \hline\hline
    L1 W587 Batch norm. & & 0.01 & $\mathbf{0.3 \pm 0.14}$ & $0.01 \pm 0.01$ & $\mathbf{-6.95 \pm 2.26}$ & $-0.57 \pm 0.29$ \\[1.5ex]
    L1 W587 & MNIST 50/50 & 0.001 & $0.0 \pm 0.0$ & $0.0 \pm 0.0$ & $-0.34 \pm 0.12$ & $0.02 \pm 0.07$ \\[1.5ex]
    L3 W256-512-256 Batch norm. &&  0.01 & $\mathbf{2.22 \pm 0.1}$ & $0.17 \pm 0.15$ & $\mathbf{-78.93 \pm 5.88}$ & $-5.06 \pm 3.68$ \\[1.5ex]
    L3 W256-512-256 &&  0.001 & $\mathbf{2.35 \pm 1.97}$ & $0.02 \pm 0.02$ & $\mathbf{-17.82 \pm 7.91}$ & $-0.4 \pm 0.42$ \\[1.5ex]
    L3 W256-512-256 batch norm. &&  0.001 & $\mathbf{0.39 \pm 0.1
}$ & - & $\mathbf{-16.29 \pm 6.35}$ & - \\[1.5ex]\hline
    L1 W126 Batch norm. & & 0.01 & $\mathbf{0.32 \pm 0.1}$ & $0.19 \pm 0.04$ & $\mathbf{-16.17 \pm 2.15}$ & $-2.77 \pm 0.79$\\[1.5ex]
    L1 W126 & CIFAR-10 50/50 & 0.001 & $0.0 \pm 0.0$ & $0.08 \pm 0.02$ & $-3.54 \pm 0.98$ & $0.98 \pm 0.39$\\[1.5ex]
    L1 W126 Batch norm. & (centered) & 0.001 & $0.06 \pm 0.02$ & $0.0 \pm 0.0$ & $-3.19 \pm 0.72$ & $2.46 \pm 0.54$\\[1.5ex]
    L3 W110-256-110 Batch norm. && 0.01 & $\mathbf{0.61 \pm 0.25}$ & $\mathbf{0.87 \pm 0.21}$ & $\mathbf{-29.78 \pm 2.57}$ & $\mathbf{-13.85 \pm 1.09}$\\[1.5ex]
    L3 W110-256-110 && 0.001 & $0.0 \pm 0.0$ & $1.01 \pm 0.24$ & $\mathbf{-16.22 \pm 3.32}$ & $-2.89 \pm 0.51$\\[1.5ex]
    L3 W110-256-110 Batch norm. && 0.001 & $\mathbf{0.61 \pm 0.1}$ & $1.09 \pm 0.24$ & $\mathbf{-18.22 \pm 0.95}$ & $-5.32 \pm 1.74$\\[1.5ex]\hline
    Resnet 10 && 0.001 & - & $\mathbf{0.8 \pm 0.31}$ & - & $\mathbf{-9.7 \pm 3.39}$\\[1.5ex]
    Resnet 10 &CIFAR-10 50/50& 0.0001 & - & $0.1 \pm 0.04$ & - & $-3.65 \pm 1.3$\\[1.5ex]
    Resnet 10 &(centered)& 0.0001 \textbf{SGDN} & - & $0.12 \pm 0.02$ & - & $-3.97 \pm 0.64$\\[1.5ex]
    Resnet 22 && 0.001 & - & $\mathbf{1.71 \pm 1.16}$ & - & $\mathbf{-28.32 \pm 13.28}$\\[1.5ex]
    Resnet 22 && 0.0001 & - & $0.14 \pm 0.06$ & - & $-1.07 \pm 0.69$\\[1.5ex]
    Resnet 22 && 0.0001 \textbf{SGDN} & - & $0.15 \pm 0.03$ & - & $-1.22 \pm 0.54$\\[1.5ex]
    VGG 11 &CIFAR-10 50/50& 0.0001 & $\mathbf{1.05 \pm 0.25}$ & $0.0 \pm 0.0$ & $\mathbf{-43.28 \pm 6.35}$ & $-1.21 \pm 0.66$\\[1.5ex]
    VGG 19 &(centered)& 0.0001 & - & $\mathbf{0.27 \pm 0.18}$ & - & $\mathbf{-24.56 \pm 4.95}$
    \\\hline\hline
    \end{tabular}}
    \caption{Decreasing learning rate or increasing batch size either reduces barrier or recovers LMC for different models. \textbf{SGDN} stands for training under different samples of SGD noise.}
\end{table*}

\begin{table*}[!ht]
\centering
    \begin{tabular}{|c|c|c|c|c|c|c|c|}
    \hline\hline
    Model & LR & Data & split $50/50$ & split $\mathbf{20/80}$ & split $50/50$ & split $\mathbf{20/80}$\\
    & & & barrier & barrier & $\Delta~(\%)$ & $\Delta~(\%)$\\ \hline\hline
    Resnet 10 & $10^{-3}$ & & $0.04 \pm 0.04$ & $\mathbf{2.51 \pm 0.68}$ & $-1.07 \pm 1.03$ & $\mathbf{-61.59 \pm 7.02}$\\[1.5ex]
    Resnet 10 & $10^{-4}$ & & $0.0 \pm 0.0$* & $\mathbf{1.9 \pm 0.31}$ & $-0.06 \pm 0.08$* & $\mathbf{-72.78 \pm 8.71}$\\[1.5ex]
    Resnet 22 & $10^{-3}$ & MNIST & $2.33 \pm 0.93$ & $2.85 \pm 0.97$ & $-54.07 \pm 11.45$ & $-63.68 \pm 13.48$\\[1.5ex]
    Resnet 22 & $10^{-4}$ & & $0.0 \pm 0.0$* & $\mathbf{1.53 \pm 0.72}$ & $0.0 \pm 0.03$* & $\mathbf{-44.59 \pm 13.39}$\\[1.5ex]\hline
    Resnet 10 & $10^{-3}$ & & $0.8 \pm 0.31$ & $\mathbf{3.86 \pm 0.81}$ & $-9.7 \pm 3.39$ & $\mathbf{-39.53 \pm 4.65}$\\[1.5ex]
    Resnet 10 & $10^{-4}$ & & $0.12 \pm 0.02$* & $\mathbf{1.8 \pm 0.56}$ & $-3.97 \pm 0.64$* & $\mathbf{-32.3 \pm 4.48}$\\[1.5ex]
    Resnet 22 & $10^{-3}$ & CIFAR-10 & $1.71 \pm 1.16$ & $\mathbf{4.22 \pm 2.03}$ & $-28.32 \pm 13.28$ & $\mathbf{-53.99 \pm 9.15}$\\[1.5ex]
    Resnet 22 & $10^{-4}$ & & $0.15 \pm 0.03$* & $\mathbf{1.95 \pm 0.91}$ & $-1.22 \pm 0.54$* & $\mathbf{-29.0 \pm 7.77}$\\[1.5ex]
    VGG 11 & $10^{-4}$ & & $0.0\pm0.0$ & $0.0\pm0.0$ & $-1.21\pm0.66$ & $1.6 \pm 0.98$\\[1.5ex]
    VGG 19 & $10^{-4}$ & & $\mathbf{0.27 \pm 0.18}$ & $\mathbf{0.29 \pm 0.18}$ & $\mathbf{-24.56 \pm 4.95}$ & $\mathbf{-27.64 \pm 11.28}$
    \\\hline\hline
    \end{tabular}
    \caption{
    Training two copies of the same model on imbalanced data leads to convergence to different local minima in case of Resnet and VGG19. Training was performed in large-batch regime (batch size 1024).\\
    * indicates that training was performed under different samples of \textbf{SGD noise}.}
\end{table*}

\clearpage
\begin{table*}[!ht]
\centering
    \begin{tabular}{|c|c|c|c|c|c|}
    \hline\hline
    Model & cent. data & [0,1] data & cent. data & [0,1] data\\
    & barrier & barrier & $\Delta~(\%)$ & $\Delta~(\%)$\\ \hline\hline
    \begin{tabular}{c}
        MLP L1\\
        B32
    \end{tabular}& $0.0\pm0.0$ & $\mathbf{0.4\pm0.12}$ & $-3.54\pm0.98$ & $\mathbf{-14.88\pm2.74}$\\[1.5ex]\hline
    \begin{tabular}{c}
        MLP L1\\
        \textbf{Batch. Norm.}, B32
    \end{tabular}& $0.06\pm0.02$ & $\mathit{0.05\pm0.03}$ & $-3.19\pm0.72$ & $\mathit{-1.62\pm1.33}$\\[1.5ex]\hline
    \begin{tabular}{c}
        MLP L1\\
        B1024
    \end{tabular}& $0.08\pm0.02$ & $\mathbf{0.74\pm0.43}$ & $0.98\pm0.39$ & $\mathbf{-15.69\pm6.63}$\\[1.5ex]\hline
    \begin{tabular}{c}
        MLP L1\\
        \textbf{Batch. Norm.}, B1024
    \end{tabular}& $0.0\pm0.0$ & $\mathit{0.0\pm0.0}$ & $2.46\pm0.54$ & $\mathit{3.49\pm0.97}$\\[1.5ex]\hline
    \begin{tabular}{c}
        MLP L3\\
        B1024
    \end{tabular}& $1.01\pm0.24$ & $\mathbf{2.6\pm0.51}$ & $-2.89\pm0.51$ & $\mathbf{-14.9\pm2.32}$\\[1.5ex]\hline
    \begin{tabular}{c}
        MLP L3\\
        \textbf{Batch. Norm.}, B1024
    \end{tabular}& $1.09\pm0.24$ & $\mathit{0.95\pm0.2}$ & $-5.32\pm1.74$ & $\mathit{-4.51\pm1.31}$\\[1.5ex]\hline
    Resnet 10, B1024 & $0.1\pm0.04$ & $0.08\pm0.03$ & $-3.65\pm1.3$ & $-3.01\pm1.42$\\[1.5ex]\hline
    VGG 11, B1024 & $0.0\pm0.0$ & $0.0\pm0.0$ & $-1.21\pm0.66$ & $-1.53\pm0.45$
    \\\hline\hline
    \end{tabular}
    \caption{LMC in wide and deep MLPs can be "broken by" normalizing image intensity to the [0,1]-range and "recovered by" adding batch normalization layer(s). Resnet and VGG models employ batch normalization layers and show stability of training dynamics for both centered and [0,1] normalized data with barrier nearly zero and accuracy drop up to $4\%$. Models were trained with learning rate of $10^{-3}$ (MLPs) and $10^{-4}$ (ResNet, VGG) in large batch size regime. Notation: MLP L1 - MLP with one hidden layer, B32 - training batch size 32.}
\end{table*}

\newpage

\end{document}